\documentclass[letterpaper, 10 pt, conference]{ieeeconf}  

\IEEEoverridecommandlockouts                              

\usepackage{etoolbox} 
\usepackage[utf8]{inputenc} 
\usepackage[T1]{fontenc}    
\usepackage{fonttable}
\usepackage[english]{babel} 
\usepackage[protrusion=true,expansion=true]{microtype}
\usepackage{comment}
\usepackage{cancel}
\usepackage{csquotes}
\usepackage{listings}
\usepackage{nameref}
\usepackage{paralist}
\usepackage{xspace}
\usepackage{setspace}
\usepackage{makeidx}
\usepackage{pdfpages}
\usepackage{hyperref}

\usepackage{times}
\usepackage{microtype}

\usepackage{mathtools}
\usepackage[
  separate-uncertainty = true,
  multi-part-units = repeat
]{siunitx}
\usepackage{adjustbox} 

\usepackage{float}
\usepackage{stfloats}

\usepackage{placeins}
\usepackage{subcaption}
\usepackage[font=small,labelfont=bf]{caption}
\usepackage{lscape}      

\usepackage{algorithm}
\usepackage[noend]{algpseudocode}

\usepackage{array} 
\usepackage{tabularx}
\usepackage{multirow}
\usepackage{multicol}
\usepackage{booktabs}   
\usepackage{tablefootnote}

\usepackage{paralist}
\let\labelindent\relax 
\usepackage{enumitem}
\setlist[itemize]{noitemsep,leftmargin=*,topsep=0in}
\setlist[enumerate]{noitemsep,leftmargin=*,topsep=0in}

\usepackage{url}            
\usepackage{color,colortbl} 
\usepackage{soul}

\usepackage[capitalize, noabbrev]{cleveref}

\hypersetup{pagebackref=false,breaklinks=true,colorlinks,urlcolor=blue,citecolor=blue,linkcolor=blue,bookmarks=false}

\usepackage{blindtext}
\usepackage{bbm}

\usepackage{lipsum}

\usepackage{titlesec}
\titlespacing{\subsection}{0pt}{0.3\baselineskip}{0.1\baselineskip}

\definecolor{Gray}{rgb}{1.0, 0.95, 0.85}  
\definecolor{Gray1}{rgb}{0.95, 0.95, 1.0}   
\definecolor{Gray2}{rgb}{0.95, 0.93, 0.82}

\def\eqref#1{equation~\ref{#1}}

\def\1{\bm{1}}

\DeclareMathAlphabet{\mathsfit}{\encodingdefault}{\sfdefault}{m}{sl}
\SetMathAlphabet{\mathsfit}{bold}{\encodingdefault}{\sfdefault}{bx}{n}

\DeclareMathOperator*{\argmax}{arg\,max}

\usepackage[
backend=biber,
style=ieee,
natbib=true,
maxbibnames=10, 
minbibnames=5, 
sorting=none
]{biblatex}
\definecolor{ishikaColor}{rgb}{1,0,0} 
\definecolor{valtsColor}{rgb}{0,1,0} 
\definecolor{stanColor}{rgb}{0,0,1} 
\definecolor{animeshColor}{rgb}{1,0.5,0} 
\definecolor{ankitColor}{rgb}{0.5,0,0.5} 

\title{\LARGE \bf
OG-VLA: Orthographic Image Generation for 3D-Aware Vision-Language Action Model
}

\author{
Ishika Singh$^{1}$, %
Ankit Goyal$^{2}$, %
Stan Birchfield$^{2}$, %
Dieter Fox$^{2}$, %
Animesh Garg$^{2}$, %
Valts Blukis$^{2}$ 
\thanks{Correspondence to: {\tt\small ishikasi@usc.edu}}
\thanks{This work was done while IS was an intern at NVIDIA}
\thanks{$^{1}$University of Southern California, $^{2}$NVIDIA}%
}
\newcommand{\ours}{\text{OG-VLA}}

\begin{document}

\maketitle
\thispagestyle{empty}
\pagestyle{empty}

\begin{abstract}

We introduce \ours, a novel architecture and learning framework that combines the generalization strengths of Vision Language Action models (VLAs) with the robustness of 3D-aware policies. 
We address the challenge of mapping natural language instructions and one or more RGBD observations to quasi-static robot actions. 
3D-aware robot policies achieve state-of-the-art performance on precise robot manipulation tasks, but struggle with generalization to unseen instructions, scenes, and objects. 
On the other hand, VLAs excel at generalizing across instructions and scenes, but can be sensitive to camera and robot pose variations. 
We leverage prior knowledge embedded in language and vision foundation models to improve generalization of 3D-aware keyframe policies. 
\ours\ unprojects input observations from diverse views into a point cloud which is then rendered from canonical orthographic views, ensuring input view invariance and consistency between input and output spaces. 
These canonical views are processed with a vision backbone, a Large Language Model (LLM), and an image diffusion model to generate images that encode the next position and orientation of the end-effector on the input scene. 
Evaluations on the \textsc{Arnold} and \textsc{Colosseum} benchmarks demonstrate state-of-the-art generalization to unseen environments, with over 40\% relative improvements while maintaining robust performance in seen settings. 
We also show real-world adaption in 3 to 5 demonstrations along with strong generalization. 
Videos, additional details, and resources at \href{https://og-vla.github.io/}{\textcolor{orange}{https://og-vla.github.io}}

\end{abstract}

\section{Introduction}

We study the problem of mapping natural language instructions and one or more posed RGBD observations to robot actions, with specific focus on quasi-static manipulation tasks that can be decomposed into a sequence of end-effector keyframes. This category encompasses a wide variety of tasks such as pick-and-place, opening/closing doors and containers, manipulating buttons, valves, switches, and more. Building robust policies that solve such tasks in \emph{unseen} environments remains an open challenge that could enable numerous industrial and household applications, from cleaning and sorting robots to machine tending.


VLAs have recently demonstrated successful generalization to concepts unseen in the robotics training data~\cite{driess2023palm,brohan2023rt2,kim24openvla,pertsch2025pi0fast}, such as manipulating novel objects based on language instructions.
While achieving generalization breakthroughs, they require massive training datasets~\cite{kim24openvla, openxembodiment,pertsch2025pi0fast} and typically accept a single RGB view input. As a result, despite the large amount of training data, the resulting systems remain sensitive to variations in camera and robot poses, which hurt their adaptability to new applications~\cite{hamster2025}.
They also lack explicit visual reasoning when predicting actions (often as LLM tokens), constraining their ability to perform precise, generalizable 3D spatial reasoning.


In contrast,
3D-aware keyframe based policies learn effectively from few demonstrations and generalize well to novel camera poses and object placements~\cite{shridhar2022peract,goyal2023rvt,goyal2024rvt,gervet2023act3d} as confirmed in the camera perturbation evaluation study in~\cite{2024colosseum}.
This success stems from a 3D scene representation within the model, such as a voxel map~\cite{shridhar2022peract}, canonical orthographic views~\cite{goyal2023rvt}, or point clouds~\cite{gervet2023act3d}.
However, unlike VLAs, these systems overfit to the training scenes and objects, failing to accept instructions that refer to new, previously unseen objects.

\begin{figure*}[h]
    \centering
    \includegraphics[trim=0 8.6cm 0 0, clip, width=\textwidth]{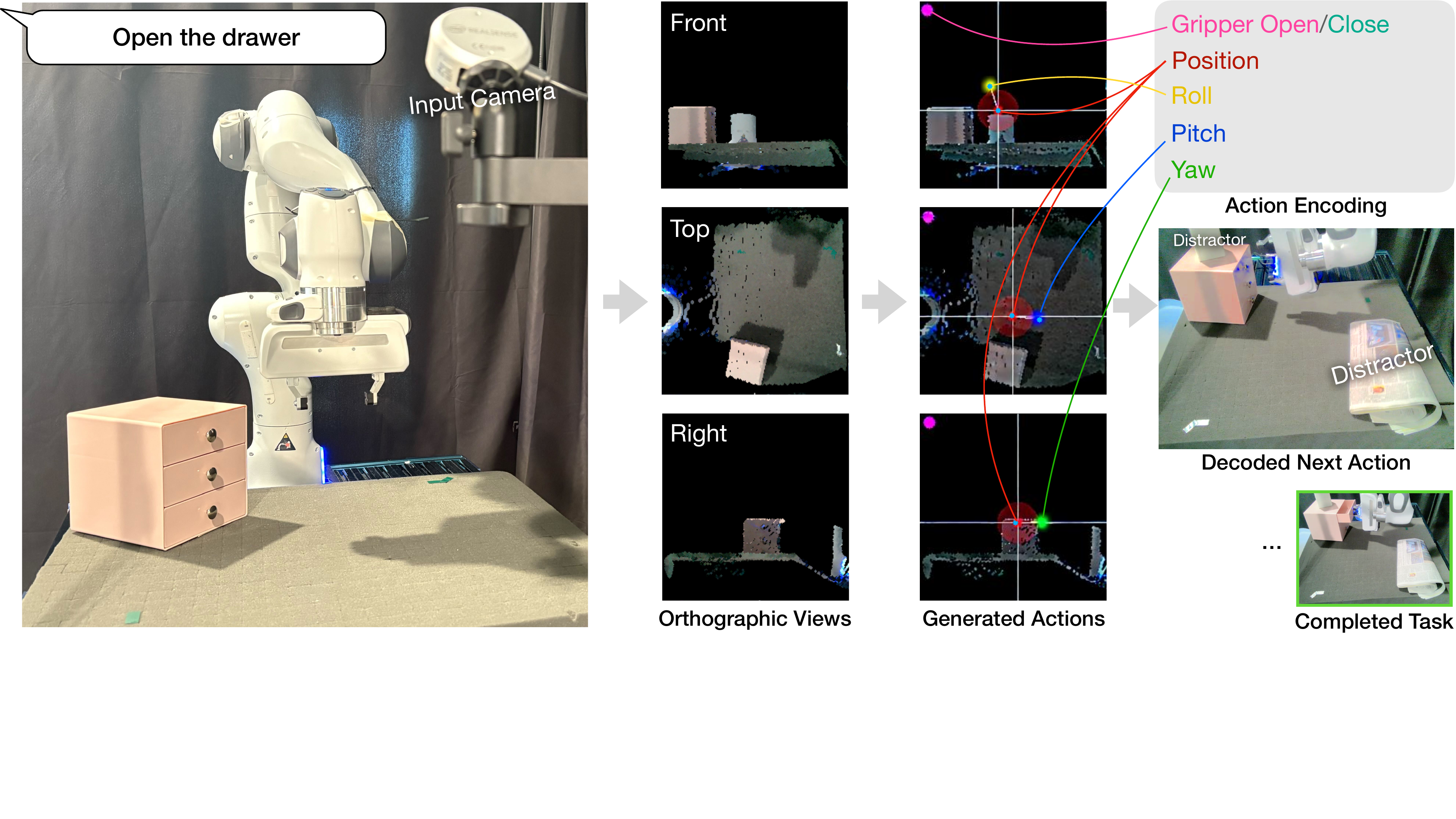}
    \caption{\textbf{\ours\ illustration.} \ours\ represents robot end-effector keyframes with easy-to-decode annotations on orthographic images in a set of canonical views. This output encoding enables action prediction via image generation, and using canonical views achieves invariance to input camera poses. The red hotspot is the predicted end-effector position in each image.
    In this example, the hotspot is indicating the 3D point for approaching the drawer handle to open it. The yellow, blue and green hotspots work in tandem with the red hotspot to encode the three axes of end-effector orientation. The color of the hotspot on the top-left encodes the gripper open/close state. The system is robust to distractors and changing lighting conditions. 
    }
    \label{fig:teaser}
\end{figure*}

We propose \ours{} (Orthographic-image Generation Vision-Language-Action model), a novel robot policy architecture that combines the generalization strengths of VLAs with robustness of 3D-aware policies.
\ours{} uses LLMs and image generation to map one or more posed RGBD images and language to 6-DOF end-effector pose keyframes in a sequence.
The system comprises four components: a point cloud renderer that renders a scene reconstruction to canonical orthographic views, a vision backbone that encodes these views into visual embeddings, an LLM that predicts action tokens, and an image diffusion model that decodes these action tokens to predict actions on each of the orthographic views through image generation, which we decode to final 3D poses.
The LLM and image diffusion models are trained end-to-end, so that they work together to produce consistent and precise predictions required for robotic manipulation.
Figure~\ref{fig:teaser} illustrates our method.

We conduct simulation and real-robot experiments.
In simulation, the \textsc{Arnold}~\cite{gong2023arnold} benchmark tests generalization to unseen objects and environments, while \textsc{Colosseum}~\cite{2024colosseum} evaluates robustness to variations in camera poses, object poses, colors, and distractors.
We show significantly improved performance on generalization tests of both benchmarks, achieving state-of-the-art on \textsc{Arnold}~\cite{gong2023arnold}.
In real-world experiments, we show our method's ability to learn manipulation tasks from as few as 3 to 5 demonstrations, highlighting its suitability for kinesthetic teaching and rapid adaptation to new domains.
We also present a detailed study of our model architecture design choices, providing insights into its functioning as well as potential future improvements.

\section{Related Work}
Large Language Models have seen an explosion in research on their use-cases in robotics, such as task planning~\cite{saycan2022arxiv,singh2023progprompt,codeaspolicies2022}, reward generation for reinforcement learning~\cite{ma2023eureka,yu2023language}, and interfacing with vision models~\cite{huang2023voxposer}.
However, although these methods generalize at a high level, they assumed access to low-level skills such pick, place, open, and close, along with perception and scene representation systems.
These skills are hard to build, and their development is in line with the goals of this work.

3D-aware keyframe-based multi-task policies have shown the ability to learn complex manipulation behaviors from as little as ten demonstrations per task. Examples of these works include PerAct~\cite{shridhar2022peract} based on voxel grids, RVT~\cite{goyal2023rvt,goyal2024rvt} based on orthonormal views, and Act3D~\cite{gervet2023act3d} based visual feature point clouds as the scene representation.
These systems significantly improve upon image-based policies~\cite{r3m, mvp} in the amount of data they require and their robustness to new object placements at test-time.
However, these systems have been trained from scratch on specific tasks, robots, and environments, and struggle to generalize to new objects and scenes, or instructions.

Vision Language Action models (VLAs) leverage large-scale prior knowledge from LLMs and vision foundation models to create robot policies that generalize to new concepts and objects at test time~\cite{driess2023palm,brohan2023rt2,kim24openvla,black2024pi0,pertsch2025pi0fast, intelligence2025pi05visionlanguageactionmodelopenworld}.
However, these systems require huge amount of demonstration data to train. For example, \cite{kim24openvla} uses over 900k demos from the Open-X Embodiment dataset~\cite{padalkar2023open}. Despite the large quantity of data, the resulting systems are sensitive to changes in the 3D environment, such as different camera poses relative to the robot system.
Our system trains with significantly less data to achieve state-of-the-art performance and generalization.

Generative image and video models have been explored for robot policies as well. 
\cite{du2023videogenpolicies} explores mapping generated videos of robots back to control via inverse dynamics.
Genima~\cite{shridhar2024generative} makes this easier by drawing robot joint annotations as textured spheres  on the video, while
RT-Trajectory~\cite{gu2023rt} generates trajectory annotations on images.
In contrast, we predict annotations on 3D canonical views, which allows us to more easily solve for 3D end-effector poses even in free space, and improves generalization from few demonstrations by enabling SE(3) data augmentation.
We show that our model can do free space reasoning for tasks such as \textit{lift the bottle 20\,cm off the ground} (Figure~\ref{fig:model_figure}).
Methods that directly predict heatmaps on images have also been previously used for language-guided  navigation tasks~\cite{anderson2019chasing, blukis2019learning}.

\section{Method: Orthographic-image Generation Vision-Language-Action model}
\begin{figure*}[t]
    \centering
    \includegraphics[trim=0 5.6cm 0 0, clip, width=\textwidth]{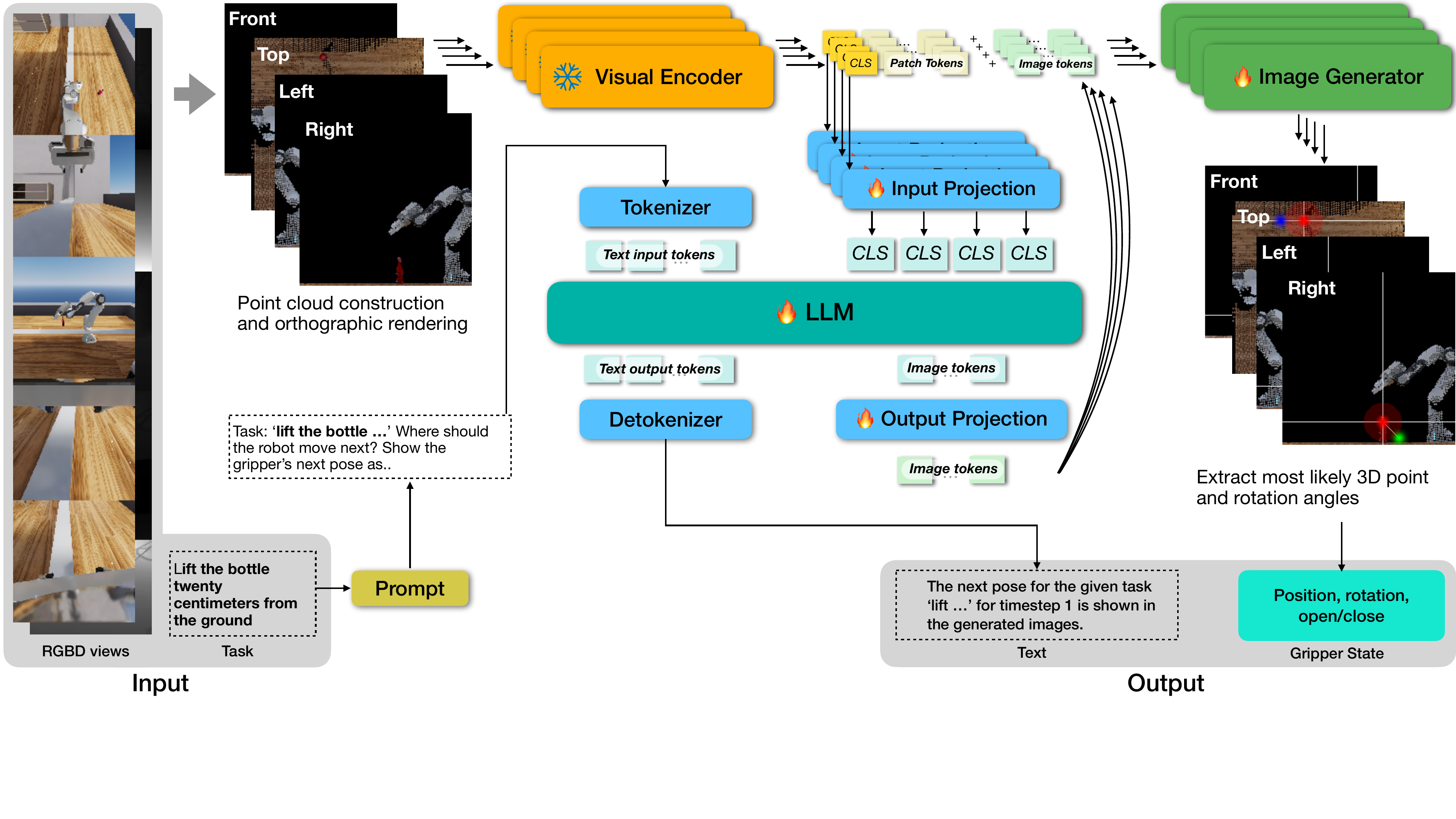}
    \caption{\textbf{Model Overview.} The input to our system is a task instruction and multiple RGB-D views of the scene. We build a point cloud from the input views and re-project it to orthographic projections from orthonormal views.
    The orthonormal views are fed into a Visual Encoder to derive a set of CLS and patch embeddings.
    CLS embeddings are projected into the LLM latent space and concatenated with a tokenized prompt that queries the next end-effector state and specifies the output format. 
    The LLM outputs image token embeddings to condition the \textsc{ImageGenerator}, which are projected to the \textsc{ImageGenerator}'s input latent space, and then concatenated with skip-connected visual features. The \textsc{ImageGenerator} generates heatmaps-one per orthographic view-indicating the next end-effector pose.
    We decode the heatmaps by interpreting them as probabilities, and inferring the most likely 3D position across all views and one rotation angle per view.
    }
    \label{fig:model_figure}
    \vspace{-1em}
\end{figure*}

At deployment time, the input to our system is a language instruction $l$ and a set of observations $O_k = \{I_{k}, D_{k}, P_{k}, K_{k}\}$, where $I_{k}$ is an RGB image, $D_{k}$ is a corresponding depth image, $P_{k}$ is the camera pose, and $K_{k}$ are the camera intrinsics, with a camera index $k$. 
The output of our system is an end-effector state $s=\langle p, \omega\rangle$, which consists of a position target $p$, rotation target $\omega$.
To complete a task, we sequentially execute our system, at each time-step using a motion planner to reach the predicted $s$, and obtain the next set of observations. Figure~\ref{fig:model_figure} shows our model architecture.

\subsection{Multi-Modal Vision and Language Model}

At the core of our system is a Large Language Model (LLM).
The LLM takes as input a sequence of input tokens (vectors) $\langle t_{1}, \dots, t_{i} \rangle$, and generates a sequence of output tokens (vectors) $\langle t_{i+1}, \dots, t_{i+j} \rangle$.
We use three types of input and output tokens: 
(1) text tokens, computed from a text tokenizer and embedding table, 
(2) input image tokens, which are either patch tokens or the image CLS token~\cite{devlin2018bert}, computed by a visual encoder, and projected to LLM space through a learned MLP input projection, and 
(3) output image (action) tokens, which we add to the LLM vocabulary and decode as a special token using an additional MLP decoder.
The output image tokens represent the next robot action.
We use an image diffusion model to decode the image tokens into actions by producing images that contain annotations that illustrate the gripper position and rotation over a set of input views of the scene.
The end-effector state $s$ is decoded from these image annotations.

\subsection{3D-Aware Reasoning with Orthogonal Orthographic Projections}
To imbue the LLM with 3D-awareness, we unproject all input camera images into a point cloud in a canonical workspace.
We then render the scene from a fixed set of views (independent of the input camera poses) before feeding them to the LLM.
This brings the input and output in the same space, and our selection of the views (orthogonal views such as `front', `top', `left', `right', rendered in orthographic mode) ensures no ambiguity in output.

\textbf{Input Reprojection to Canonical Views. }
For each camera observation $\{I_{k}, D_{k}, P_{k}, K_{k}\}$ with $N_{k}$ valid depth pixels, we compute a point cloud $C_{k} \in \mathbbm{R}^{N_{k} \times 6}$, where each contains the RGB color and 3D coordinate in a fixed reference frame.
We compute an aggregated point cloud $C = \bigcup_{k=1}^{K} C_k$ over all input cameras.
Next, we define a set of $m$ canonical cameras $\{P^{C}_{c}, K^{C}_{c}\}_{c=1,\dots,m}$, where each $P^{C}_{c}$ is a camera pose and $K^{C}_{c}$ are the intrinsics for a given orthograpic camera $c$.
We then render the point cloud to an RGB image $I^{C}_{c} \in \mathbbm{R}^{h \times w \times 3}$ seen by each orthographic camera.

Our system allows adapting the set of canonical camera parameters based on application and workspace geometry.
In this work, we use four orthographic cameras that view the workspace from front, left, right and top directions,
such that the workspace fills the camera image.
Figure~\ref{fig:model_figure} shows the input RGBD views and the resulting orthographic images.
While we use point clouds, our method may accommodate any 3D representation (e.g., neural radiance fields~\cite{mildenhall2021nerf, yu2021pixelnerf} or novel-view synthesis methods~\cite{kulhanek2022viewformer}) that supports canonical view rendering.

\textbf{LLM Input and Output. }
The canonical views $\{I^{C}_{c}\}_{c=1, \dots, m}$ constitute the visual input to our VLM system.
We process each view with a \textsc{VisualEncoder} and obtain an image embedding (CLS token) $e_{c}^{CLS}$, as well as a sequence of image patch embeddings $\langle e_{c}^{1}, \dots, e_{c}^{n} \rangle$.
Next, we apply an input projection neural network to map each CLS embedding $e_{c}^{CLS}, c \in \{1, \dots, m\}$ into a token $t_{c}^{CLS}$ compatible with the LLM input space.
Finally, the input sequence to the LLM is the following sequence of tokens: $\langle \textsc{Prompt}(l), t^{1}_{CLS}, \dots ,t^{m}_{CLS} \rangle$, where $\textsc{Prompt}(l)$ is a function that constructs a prompt for the instruction $l$ and tokenizes it in a way compatible with the LLM.




We feed the input sequence to the LLM to produce an output sequence
of format: $\langle t^{1}_{a}, t^{2}_{a}, t^{3}_{a}, t^{4}_{a}, t^{o}_{1}, t^{o}_{2}, \dots, t^{o}_{j} \rangle,$
where $t^{(\cdot)}_{a}$ are four image tokens that together represent the next action, and $t^{o}_{(\cdot)}$ are accompanying text tokens as a response to the input prompt.
We show the abbreviated prompt in Figure~\ref{fig:model_figure}. Full prompt and text response is provided in the Appendix on project website.

\textbf{Image Token Decoding for Action Prediction. }
Successful robot manipulation, such as picking objects or grabbing drawer handles, requires very precise end-effector position predictions. Although the LLM can output gripper poses in text with generally close predictions, we find them to lack the precision needed for practical applications.
We thus propose decoding the image tokens to  action annotations over each of the canonical images.
From these annotations, we can infer the next 3D gripper position and orientation by decoding and aggregating the image annotations in 3D.

First, we project each of the output image tokens $t^{i}_{a}$ back to the visual embedding space with an output projection layer to obtain a set of embeddings ${e^{i}_{a} \; ; \; i \in \{1,\dots,4\}}$.
Next, we use these embeddings as well as the output of the \textsc{VisualEncoder} (all patch embeddings $\langle e_{c}^{1}, \dots, e_{c}^{n} \rangle$, as well as the CLS token $e_{c}^{CLS}$) to condition an \textsc{ImageGenerator} network for each orthographic camera $c \in \{1,\dots,m\}$.
The network outputs an RGB image with action predictions overlaid on the input canonical views:
\begin{align}
    H^{c} = &\textsc{ ImageGenerator}(e^{i}_{a}, e_{c}^{x}), \nonumber \\
    i \in &\{1,\dots,4\} \quad c \in \{1,\dots,m\} \quad x \in \{CLS, 1, \dots,256\} 
\end{align}
where $H_{c} \in \{\mathbbm{R}^{h,w,3}\}$ is a reconstruction of the input canonical image $I^{C}_{c}$ with overlaid annotations that encode the gripper position and orientation. Each image is aligned with one of the $m$ canonical input views $c$, as shown in Figure~\ref{fig:model_figure}.
While any image generator can be used, we use StableDiffusion 1.5~\cite{rombach2021stablediffusion}.
$e^{i}_{a}$ acts as the textual conditioning and $e_{c}^{x}$ as the visual conditioning for the \textsc{ImageGenerator}. 
In practice, we use $e^{i}_{a} + \text{CLIP}(\textsc{Prompt}(l))$ for textual conditioning, essentially adding a residual skip connection from a direct text encoding, a design choice informed by empirical experiments. CLIP is the textual encoder used for pretraining the StableDiffusion 1.5 model.

\textbf{Extracting 3D Position and Rotation from Generated Images. }
We generate gripper position and Euler rotation as Gaussian distributions overlaid in different color channels on the input orthographic views. We infer a 3D position $p^{hm}$ that best explains the predictions in each of the canonical views by solving the optimization problem:
\begin{align}
    p^{hm} = \argmax_{p} \prod_{c=1,\dots,m} ( H_{c}[\mathcal{C}(p, P^{C}_{c}, K^{C}_{c})] + \epsilon)
\label{eq:argmax3d}
\end{align}
%
where $\mathcal{C}$ projects the 3D point $p$ to 2D image coordinates w.r.t.\ the orthographic camera $c$.
The square brackets represent a 2D pixel-wise indexing operation with interpolation to support sub-pixel coordinates. $\epsilon$ is a small value we add to allow decoding in situations where one of the heatmaps is zero for all 3D points.

We predict Euler rotations on the images such that the rotation along an axis is overlaid on the canonical view along that axis using 3 different colors. 
We present x-axis rotation on the front view, y-axis rotation on left and right views, and z-axis rotation on the top view as shown in Figure~\ref{fig:model_figure}.
Rotations are overlaid as Gaussian distributions at 30 pixel radius with reference to a horizontal line drawn from the translation point towards the right of the image. 
To decode rotation, we first extract the pixel location of the most likely rotation ($r^c_x, r^c_y$) and translation point ($p^c_x, p^c_y$) from the Gaussian distributions using a filtering operation for each view, and then compute the rotation angles using the arctangent function. The gripper's open/close state is encoded with binary colored hotspots on the top-left of the image, as shown in Figure~\ref{fig:teaser}.

\subsection{Training and Implementation Details}
Our architecture builds on X-VILA~\cite{ye2024xvila}, a multi-modal chat model supporting language, visual, video, and audio modalities. \ours\ is initialized from X-VILA’s pretrained weights to leverage its any-to-any modality alignment, focusing on text-image input to text-image output alignment. We leave study of enriching human-robot interaction using the other modalities to future work.
We train \ours\ using DeepSpeed~\cite{rasley2020deepspeed}. We freeze the visual encoder and tune the $\textsc{LLM}$, input and output projection networks, and $\textsc{ImageGenerator}$ with end-to-end gradient flow. Following X-VILA, we use ImageBind~\cite{girdhar2023imagebind} as the visual encoder, linear layers for input and output projections, and Stable Diffusion 1.5~\cite{sd15} for the \textsc{ImageGenerator}. The LLM is based on Vicuna-7b v1.5 architecture. ImageBind encodes images into 256 patches (16×16) with a CLS token. Each training sample includes a natural language instruction $l$, visual observations ${I_{k}, D_{k}, P_{k}, K_{k}}$, and ground-truth gripper state $\hat{s}$. 
Following prior work~\cite{shridhar2022peract, goyal2023rvt}, we augment each keyframe with $N$ SE(3)-transformed purterbations: translation in [±0.1m, ±0.1m, ±0.1m], rotation in [±0°, ±0°, ±90°]. All models are trained on 8×A100 GPUs with batch size 64. All models are trained for one run, and the LLM is LoRA-finetuned.
Inference is done on a single A100 GPU.
For inference-time image sampling from Stable Diffusion 1.5, we use 100 steps and a guidance scale of 7.0. These parameters result in reasonable sampled image quality and avoid jitters in the sampled image from the model's latent space.

\section{Simulation Experiments}
\definecolor{Gray}{rgb}{1.0, 0.95, 0.85}  
\definecolor{Gray1}{rgb}{0.95, 0.95, 1.0}   
\definecolor{Gray2}{rgb}{0.95, 0.93, 0.82}


\begin{table*}[t!]
\centering
\resizebox{\textwidth}{!}{%
\begin{tabular}{lcccccccccccccc}
\toprule
      & Pickup & Reorient & Open & Close & Open & Close & Pour & Transfer & Overall \\ 
Model & Object & Object & Drawer & Drawer & Cabinet & Cabinet & Water & Water &  \\ 
\midrule
\textbf{6D-CLIPort~\cite{zheng2022vlmbench}}  & \,\,6.7 & 0.0 & 0.0 & 0.0 & 0.0 & 0.0 & 0.0 & 0.0 & 0.8 \\
\rowcolor{Gray}
\quad -Novel Object        & \,\,8.4 & 0.0 & 0.0 & 0.0 & 0.0 & 0.0 & 0.0 & 0.0 & 1.0 \\
\rowcolor{Gray1}
\quad -Novel Scene         & 10.4 & 0.0 & 0.0 & 0.0 & 0.0 & 1.3 & 0.0 & 0.0 & 1.5 \\
\rowcolor{Gray2}
\quad -Novel State         & \,\,0.0 & 0.0 & 0.0 & 0.8 & 0.0 & 0.0 & 0.0 & 0.0 & 0.1 \\

\midrule
\textbf{PerAct~\cite{shridhar2022peract}}  & 83.3 {\tiny$\pm$  2.4} & \textbf{16.7} {\tiny$\pm$  6.2} & 30.0 {\tiny$\pm$ 10.8} & 31.7 {\tiny$\pm$  8.5} & \textbf{25.0} {\tiny$\pm$  0.0} & \textbf{30.0} {\tiny$\pm$  0.0} & \textbf{36.7} {\tiny$\pm$  6.2} & 18.3 {\tiny$\pm$  2.4} & 34.0 {\tiny$\pm$  3.1}
 \\
\rowcolor{Gray}
\quad -Novel Object     & 75.0 {\tiny$\pm$  0.0} &  3.3 {\tiny$\pm$  2.4} &  0.0 {\tiny$\pm$  0.0} & 23.3 {\tiny$\pm$ 13.1} &  0.0 {\tiny$\pm$  0.0} &  0.0 {\tiny$\pm$  0.0} & 30.0 {\tiny$\pm$  4.1} &  1.7 {\tiny$\pm$  2.4} & 16.7 {\tiny$\pm$  2.6}
 \\
\rowcolor{Gray1}
\quad -Novel Scene      & \textbf{75.0} {\tiny$\pm$  4.1} & \textbf{13.3} {\tiny$\pm$  2.4} & 13.3 {\tiny$\pm$  9.4} & 30.0 {\tiny$\pm$ 14.1} &  0.0 {\tiny$\pm$  0.0} &  \textbf{6.7} {\tiny$\pm$  2.4} & \textbf{26.7} {\tiny$\pm$  6.2} &  3.3 {\tiny$\pm$  2.4} & 21.0 {\tiny$\pm$  3.1}
 \\
\rowcolor{Gray2}
\quad -Novel State      & \textbf{16.7} {\tiny$\pm$  2.4} &  1.7 {\tiny$\pm$  2.4} &  5.0 {\tiny$\pm$  0.0} & 11.7 {\tiny$\pm$  6.2} &  0.0 {\tiny$\pm$  0.0} &  0.0 {\tiny$\pm$  0.0} &  5.0 {\tiny$\pm$  0.0} & 11.7 {\tiny$\pm$  2.4} &  6.5 {\tiny$\pm$  1.2}
 \\
\midrule
\textbf{\ours @30k}  & 86.7{\tiny$\pm$2.9} & 15.0{\tiny$\pm$8.7} & 38.3{\tiny$\pm$2.9} & 51.7{\tiny$\pm$2.9} & 0.0{\tiny$\pm$0.0} & 16.7{\tiny$\pm$2.9} & 25.0{\tiny$\pm$5.0}  & 16.7{\tiny$\pm$7.6} & 31.2{\tiny$\pm$2.9}  \\
\rowcolor{Gray}
\quad -Novel Object & \textbf{85.0}{\tiny$\pm$5.0}  & 0.0{\tiny$\pm$0.0}   & \textbf{1.7}{\tiny$\pm$2.9} & 55.0{\tiny$\pm$13.2} & \textbf{1.7}{\tiny$\pm$2.9} & \textbf{5.0}{\tiny$\pm$5.0}  & 18.3{\tiny$\pm$2.9}  & 6.7{\tiny$\pm$7.6}  & 21.7{\tiny$\pm$0.7} \\
\rowcolor{Gray1}
\quad -Novel Scene & 70.0{\tiny$\pm$2.9} & 1.7{\tiny$\pm$2.8} & 26.7{\tiny$\pm$11.5} & 36.7{\tiny$\pm$5.8}  & \textbf{1.7}{\tiny$\pm$2.9} & 1.7{\tiny$\pm$2.9} & 16.7{\tiny$\pm$11.5} & 8.3{\tiny$\pm$2.9} & 20.8{\tiny$\pm$1.3} \\
\rowcolor{Gray2}
\quad -Novel State & 0.0{\tiny$\pm$0.0}   & \textbf{13.3}{\tiny$\pm$7.6}   & 13.3{\tiny$\pm$2.9} &  \textbf{20.0}{\tiny$\pm$0.0} & 0.0{\tiny$\pm$0.0}     & 0.0{\tiny$\pm$0.0} & \textbf{8.3}{\tiny$\pm$7.6}   & 13.3{\tiny$\pm$2.9}    &  8.5{\tiny$\pm$1.9} \\
\midrule
\textbf{\ours @100k}  & \textbf{88.3} {\tiny$\pm$  2.4} & \textbf{16.7} {\tiny$\pm$  9.4} & \textbf{48.3} {\tiny$\pm$  2.4} & \textbf{56.7} {\tiny$\pm$  2.4} &  6.7 {\tiny$\pm$  4.7} & 23.3 {\tiny$\pm$ 16.5} & 33.3 {\tiny$\pm$  6.2} & \textbf{28.3} {\tiny$\pm$  2.4} & \textbf{37.7} {\tiny$\pm$  0.6}
 \\
\rowcolor{Gray}
\quad -Novel Object       & 65.0 {\tiny$\pm$  8.2} & \textbf{15.0} {\tiny$\pm$  4.1} &  \textbf{1.7} {\tiny$\pm$  2.4} & \textbf{58.3} {\tiny$\pm$ 12.5} &  0.0 {\tiny$\pm$  0.0} & \textbf{5.0} {\tiny$\pm$  4.1} & \textbf{45.0} {\tiny$\pm$  8.2} &  \textbf{8.3} {\tiny$\pm$  4.7} & \textbf{24.8} {\tiny$\pm$  1.2}
 \\
\rowcolor{Gray1}
\quad -Novel Scene       & \textbf{75.0} {\tiny$\pm$  7.1} & \textbf{13.3} {\tiny$\pm$  8.5} & \textbf{31.7} {\tiny$\pm$  4.7} & \textbf{51.7} {\tiny$\pm$  2.4} &  \textbf{1.7} {\tiny$\pm$  2.4} &  5.0 {\tiny$\pm$  4.1} & \textbf{26.7} {\tiny$\pm$  2.4} & \textbf{25.0} {\tiny$\pm$  7.1} & \textbf{28.8} {\tiny$\pm$  0.5}
 \\
\rowcolor{Gray2}
\quad -Novel State      &  0.0 {\tiny$\pm$  0.0} & \textbf{13.3} {\tiny$\pm$  2.4} & \textbf{25.0} {\tiny$\pm$  7.1} & 15.0 {\tiny$\pm$  4.1} &  0.0 {\tiny$\pm$  0.0} &  0.0 {\tiny$\pm$  0.0} &  6.7 {\tiny$\pm$  4.7} & \textbf{20.0} {\tiny$\pm$  7.1} & \textbf{10.0} {\tiny$\pm$  0.9}
 \\

\bottomrule
\end{tabular}
}
\caption{\textbf{Success rate on ARNOLD~\cite{gong2023arnold}.} Success rates for 8 tasks and 4 test splits are shown for 2 baseline models and our model at specified training iterations (30k and 100k). 
The first row for each model is the Novel Pose split (the Test set in \textsc{Arnold}).
The numbers in bold represent the best system for each task and test split.
}
\label{tab:results1}
\end{table*}

\subsection{Benchmarks and Datasets}

We evaluate our method on the \textsc{Arnold}~\cite{gong2023arnold} and \textsc{Colosseum}~\cite{2024colosseum} benchmarks, which test language-grounded robot task learning in realistic 3D scenes with emphasis on testing generalization. \textsc{Arnold} uses five input cameras (front, base, left, wrist top, wrist bottom) and includes eight language-conditioned tasks (see Table~\ref{tab:results1}), each with four generalization test splits: (1) Novel Pose (held-out object/robot placements), (2) Novel Object (unseen objects), (3) Novel Scene (unseen scenes with seen objects), and (4) Novel State (unseen goal states). \textsc{Arnold} tasks follow a two-keyframe format—grasping and manipulating (e.g., pull drawer to 50\% open)—and do not require gripper state prediction. These tasks demand both object pose and free space reasoning.
\textsc{Colosseum} has four cameras (front, left shoulder, right shoulder, wrist) and features 20 language-conditioned tabletop tasks (e.g., close box, empty dishwasher) with 2–13 gripper keyframes (average 6), requiring gripper open/close prediction. It evaluates generalization via the \textit{all perturbation} test set, which simultaneously alters object/table/background appearance, lighting, camera pose, and adds distractors.

We train on \textsc{Arnold}’s training split with 8 tasks, \textasciitilde500 demos/task, and 2 keyframes/demo (7100 keyframes). We separately train on \textsc{Colosseum} training split with 100 demos/task. For SE(3) augmentations, we set $N=10$ for \textsc{Arnold} (\textasciitilde70k training samples) and $N=5$ for \textsc{Colosseum} (\textasciitilde1M training samples).
All \textsc{Arnold} models are trained for 30k iterations and the final result is reported at 30k and 100k iterations (30k takes 1.5 days, and 100k takes 5 days to train). We train on \textsc{Colosseum} for 250k iterations due to the larger training data size. We noticed that further training beyond 250k iterations degrades image generation quality. We conduct ablation studies at 30k on \textsc{Arnold} due to compute constraints. 

Each \textsc{Arnold} test split includes 20 episodes; \textsc{Colosseum} uses 25.  We evaluate each \textsc{Arnold} model over 3 runs to account for simulator's motion planner noise during keypoint execution, and we report means and standard deviations. Baseline models were trained by the respective benchmark authors. We re-run PerAct~\cite{shridhar2022peract} evaluation for 3 runs on \textsc{Arnold} for fair comparison, and report 6D-CLIPort~\cite{zheng2022vlmbench} results from \textsc{Arnold}~\cite{gong2023arnold} due to lack of checkpoint release.

For state-of-the-art (SotA) VLA comparisons, we include $\pi_0$-FAST~\cite{pertsch2025pi0fast} and $\pi_{0.5}$~\cite{intelligence2025pi05visionlanguageactionmodelopenworld} baselines, reported in Table~\ref{tab:results6}.
We compare single task VLA policies due to compute constraints. 
We finetune provided checkpoints pretrained on 10k+ hours of data across various robot setups with actions in both joint and end-effector continuous control spaces.
All models are trained and evaluated on Pickup Object task with 600 training and 20 test episodes, batch size 64, and 30k iterations.
Each $\pi$ model was finetuned on 4xL40 GPUs (LoRA: 0.5 days; Full: 2 days). 
At inference, we execute first half of the predicted action chunk (5 actions) for temporal consistency, inferring 80 times per episode to allow 400 environment steps.

\subsection{Simulation Results}

\subsubsection{\textsc{Arnold}}
We present our model results in Table~\ref{tab:results1}. 
\ours\ outperforms the baseline (PerAct) for most of the tasks' Novel Pose split (seen objects and scenes), with a task-averaged relative improvement of 10.8\%.
We present our result at 30k and 100k iterations.
\ours\ improves significantly across generalization splits (Novel Object, Scene, and State), with relative increases of 20.0\% and 46.5\% respectively at 30k and 100k iterations in overall success rates.
This demonstrates the effectiveness of our method in adapting pretrained visual and textual priors to robotics applications.

\begin{table}[h!]
    \centering
    \resizebox{1.0\linewidth}{!} {
    \begin{tabular}{lcccc}
        \toprule
              Model $\rightarrow$ & $\pi_0$-FAST & $\pi_0$-FAST & $\pi_{0.5}$ & OG-VLA \\ 
              & (LoRA) & (Full) & (Full) & (LoRA) \\
        \midrule
        Pickup Object   & 25.0 & 35.0 & 0.0 & \textbf{95.0}   \\
        \rowcolor{Gray}
        \quad - Novel Object  & 25.0 & 50.0 & 5.0 & \textbf{90.0}   \\
        \rowcolor{Gray1}
        \quad - Novel Scene & 20.0 & 35.0 & 0.0 & \textbf{80.0}   \\
        \rowcolor{Gray2}
        \quad - Novel State & 10.0 & \textbf{30.0} & 0.0 & \,\,\,0.0   \\
        \bottomrule
    \end{tabular}
    }
    \caption{\textbf{Comparison with VLAs (\%).} Success rates are reported for Pickup Object \textsc{Arnold} task on Novel Pose, Novel Object, Novel Scene and Novel State test splits, averaged over 20 episodes each.
    }
    \label{tab:results6}
\end{table} \vspace{-0.5cm}
\paragraph{\textsc{Comparison with VLAs}}

SotA VLA baselines significantly underperform \ours\ even on the simplest \textsc{Arnold} task (Table~\ref{tab:results6}).
$\pi_0$-FAST learns the task, likely due to \textsc{Arnold} data's action frequency being more suitable for its discrete frequency-space action tokenizer, unlike for the flow-matching decoder used for $\pi_{0.5}$.
$\pi_0$-FAST shows promise on Novel State, reflecting its preserved language understanding, whereas \ours\ struggles comparatively on this task, likely due to the lack of similar pretraining. Co-training with external datasets can improve OG-VLA, as show in prior works~\cite{yuan2024robopoint}.
These results confirm that injecting 3D understanding in VLAs significantly improves them while preserving their generalization capabilities. 


\begin{table}[h!]
    \centering
    \resizebox{1.0\linewidth}{!} {
    \begin{tabular}{lccc}
        \toprule
             Model Latency & step (s)$\downarrow$ & \#steps/episode & episode (s)$\downarrow$ \\
        \midrule
        $\pi_0$-FAST (LoRA)   & 0.4 & 80 & 102.5 (42.1$^*$)   \\
        $\pi_0$-FAST    & 0.4 & 80 & 103.1 (39.1$^*$)   \\
        $\pi_{0.5}$       & 0.2 & 80 & 70.3\,\,\,\,\,\,\,\,\,\,\,\,\,\,\,\,   \\
        PerAct          & 0.2 & \,\,\,2 & \,\,\,1.5\,\,\,\,\,\,\,\,\,\,\,\,\,\,\,\,   \\
        OG-VLA (LoRA)   & 4.5 & \,\,\,2 & 10.2\,\,\,\,\,\,\,\,\,\,\,\,\,\,\,\,   \\
        \bottomrule
    \end{tabular}
    }
    \caption{\textbf{Model Latency (seconds).} Reported for per inference step, number of inference steps required per episode by each model, and finally overall episode latency with simulation execution for Pickup Object \textsc{Arnold} task, averaged over 20 episodes. $^*$indicates model inference time reported only on successful episodes.
    }
    \label{tab:results7}
\end{table} \vspace{-0.5cm}
\paragraph{\textsc{Model Latency}}
Table~\ref{tab:results7} reports inference latency for both the baseline models and \ours. While per-step inference latency is higher for \ours\ due to multiple encoder–decoder modules and data processing involved during prediction, our model operates as a keyframe policy and thus requires only a few inference steps per episode, similar to other 3D keyframe policies~\cite{shridhar2022peract, goyal2023rvt}.
As a result, overall episode inference time is much lower than VLA baselines, since the motion planner adds negligible cost compared to continuous control predictions in VLAs. 
However, \ours\ remains slower than PerAct. We defer this optimization to future work.


\begin{figure}[h!]
     \centering
    \includegraphics[width=0.8\linewidth]{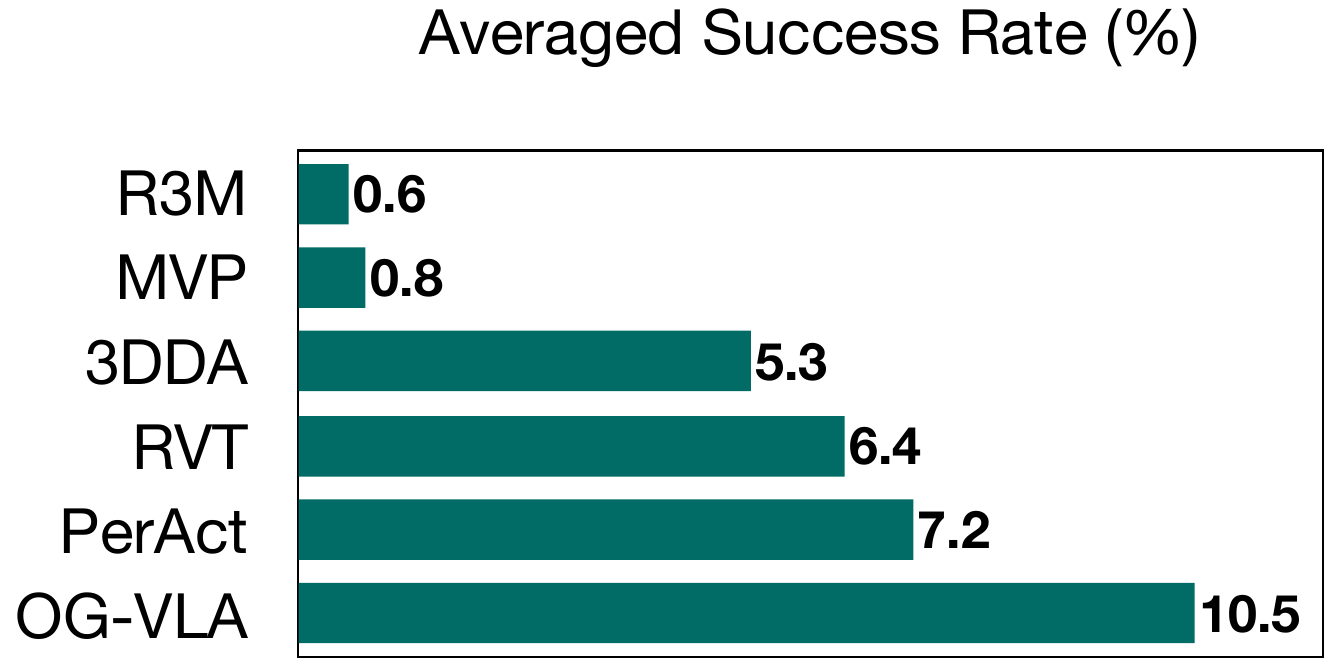}
    \caption{\textbf{Evaluation on the COLOSSEUM benchmark~\cite{2024colosseum}.} Task-averaged success rate shows that \ours\ outperforms all baselines on the hardest generalization test set (\emph{all perturbation}).}
    \label{fig:colosseum}
\end{figure} 
\subsubsection{\textsc{Colosseum}}
Figure~\ref{fig:colosseum} shows task-averaged success rate on \textit{all perturbation} test set on the \textsc{Colosseum} benchmark. We compare with baselines reported in \textsc{Colosseum}, including R3M~\cite{r3m}, MVP~\cite{mvp}, 3DDA~\cite{3d_diffuser_actor}, RVT~\cite{goyal2023rvt}, and PerAct~\cite{shridhar2022peract}. 
\ours\ improves upon the baselines by a relative increase of 45.8\%.
The absolute performance remains low however (10.5\%). We observe that \textsc{Colosseum} tasks are much harder to learn due to longer sequence of keyframe predictions, leading to error accumulation for an imitation learning based method.

\subsection{Design Choice Ablations}
We conducted extensive ablations on the \textsc{Arnold} benchmark to validate our design choices. We highlight the main findings in this section, with detailed results and analysis in Appendix on project website.

\paragraph{Action Prediction Approaches} We experimented with two common alternatives to our image generation approach: (1) direct text-based action prediction and (2) adding additional action tokens to the LLM vocabulary that are decoded with MLP decoders directly to gripper states. Both alternatives failed to learn effective policies under our training conditions, likely due to insufficient visual reasoning capabilities for precise manipulation tasks with limited training data. 

\paragraph{Image Generation Modes} We compared three image generation approaches: (1) generation without scene reconstruction (black background), (2) generation with scene reconstruction (action annotations overlaid on the input image), and (3) generation with faded reconstruction (rescaled to 0-127 range, keeping colors in range 128-255 reserved for action annotations). Generation with faded reconstruction performed best overall on generalization splits (e.g. 23.8\% vs. 12.8\% without reconstruction and 20.8\% with full reconstruction on the \textsc{Arnold} novel scene split).
Generating full reconstructions complicates action decoding due to occasional color collisions. Generating black backgrounds was unstable during training,
likely due to the challenge of adapting a model of natural images towards generating plain backgrounds.

\paragraph{Architecture Components} Our ablations revealed several key insights: (1) Unlike prior work, tiling orthographic views decreased performance (24.8\% vs. 31.2\%); (2) Removing the LLM significantly reduced performance (20.0\% vs. 31.2\%), likely due to its strong priors and its role in conditioning consistent generation across views; (3) Directly bypassing the instruction to Image Generator decreased performance by 9.5\%, suggesting that LLM's image token outputs preserve crucial task information. These findings validate the importance of each component in OG-VLA.

\section{Real Robot Experiments}
\paragraph{Experimental Setup} We perform real-world experiments on a Franka Emika Panda arm mounted on a tabletop, with a single front-facing camera, as shown in Figure~\ref{fig:teaser}. We collect 3–5 demos for 4 real-world tasks—22 demos in total—with human-annotated keyframes and a motion planner~\cite{zhu2022viola-deoxys} to achieve annotated keyframes. We train both the baseline and our model on this dataset. For our model, we augment each keyframe with 10 SE(3) perturbations and finetune the Arnold-pretrained \ours@30k checkpoint for another 10k iterations with a batch size of 64.
We use $\pi_0$-FAST Full finetuning as our baseline due to its strongest performance in simulation.
We finetune it on our dataset with joint angle actions for 30k iterations with a batch size of 32, as used in most of their pre-training and finetuning experiments~\cite{pertsch2025pi0fast, black2024pi0}. During inference, the model predicts an action chunk of 10 actions; we execute the last action in the sequence for faster execution.

\paragraph{Quantitative Results} We report our results in Table~\ref{tab:results5}. Each test set success rate is averaged over 10 episodes. For novel object variation, we use unseen colored and shaped objects. For novel scene variation, we introduce distractors and change lighting, background, and table appearance.
Please refer to Appendix on project website for training and test scene details.
Our results show that \ours\ can adapt to new tasks with only 3–5 demonstrations and generalize well to novel poses, objects, and scenes.  Although we attempted to compare with $\pi_0$-FAST,  it failed at all tasks, learning only to reach the block with 30\% success.  We hypothesize that this is due to the small number of demonstrations and lack of support for SE(3) data augmentation without our 3D representation.

\begin{table}[h!]
    \centering
    \resizebox{1.0\linewidth}{!} {
    \begin{tabular}{lcccc}
        \toprule
         & Pickup & Put Object & Open  & Close \\ 
        Model & Object & in Drawer &  Drawer &  Drawer \\ 
        \midrule
        \textbf{\ours @10k}   & 100.0 & 90.0 & 60.0 & 90.0   \\
        \rowcolor{Gray}
        \quad - Novel Object  & \,\,80.0 & 70.0 & 30.0 & 50.0   \\
        \rowcolor{Gray1}
        \quad - Novel Scene & \,\,90.0 & 80.0 & 50.0 & 90.0   \\
        \bottomrule
    \end{tabular}
    }
    \caption{\textbf{Real world success rate (\%).} Success rates are reported for 4 real world tasks and novel pose, novel object, and novel scene test splits, averaged over 10 episodes each.
    }
    \label{tab:results5}
\end{table}

\begin{figure}[h!]
    \centering
    \includegraphics[width=\linewidth]{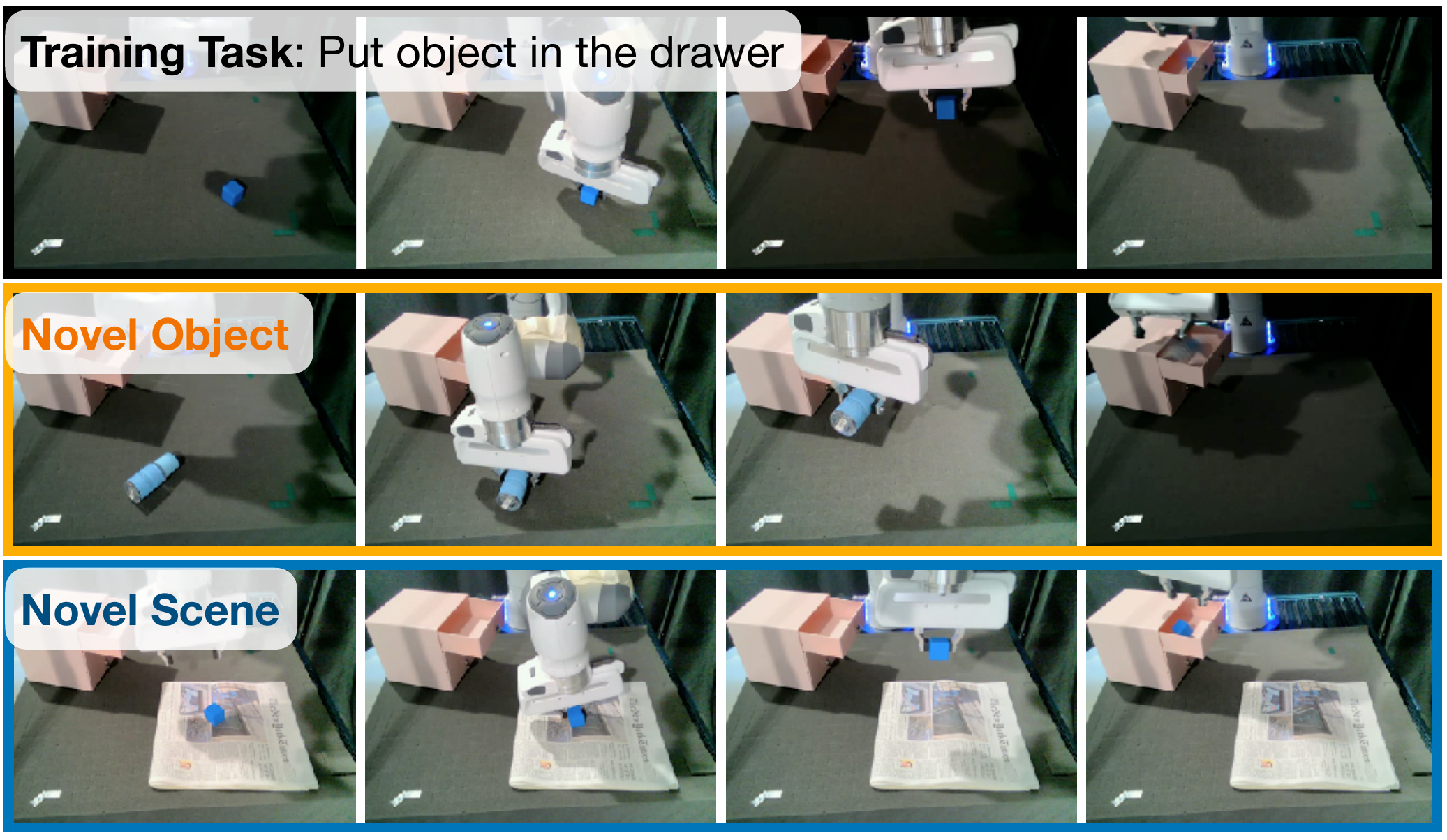}
    \caption{\textbf{Qualitative example.} Showing generalization of \ours\ to unseen scenarios.}
    \label{fig:qualitative}
\end{figure}

\vspace{-0.5cm}
\paragraph{Qualitative Results}
We show qualitative examples of real world evaluations in Figure~\ref{fig:qualitative} for the task `put object in the drawer'. For training demonstrations, we use a blue cube. During evaluation, we replace it with bottle or perturb the scene by placing a newspaper under the cube also make te scene brighter. \ours\ can generalize to manipulating differnt objects as well as to unseen scenes for a given task.

\vspace{-0.5cm}
\section{Conclusion and Limitations}
We introduced \ours, a novel architecture and learning framework that combines the generalization strengths of VLA models with the robustness of 3D-aware keyframe policies for robotic manipulation. By leveraging foundation models in language and vision, \ours\ improves generalization to unseen instructions, objects, and scenes while maintaining precise control. Our approach ensures input-view invariance by projecting one or more RGBD observations into a canonical point cloud representation, which is processed through a vision backbone, an LLM, and an image diffusion model to generate actions as images. 
\ours\ achieves state-of-the-art performance on robotic manipulation tasks, particularly on scene and object generalization tests.
Moreover, \ours\ can adapt to real-world tasks in 3-5 demonstrations with strong generalization to unseen objects and scenes. 
In practice, despite per-step inference being relatively slow, \ours\ operates efficiently as a keyframe policy that predicts actions only at a low frequency, with intermediate motion handled by a motion planner. This allows fluid execution in real-world tasks even with a single camera, highlighting the flexibility of our framework.
These results highlight the effectiveness of integrating pretrained visual and text priors into a structured 3D-aware framework for robot learning. 
Future work will explore extending \ours\ to complex long-horizon tasks, external data augmentation, co-training techniques, and model optimization for faster inference.

\ours\ has several strengths but also natural limitations.
One challenge arises from the reliance on orthographic canonical views, which, while effective in many scenarios, can be affected by severe occlusions, such as multiple objects stacked on a shelf.
Occlusions may lead to partial or incorrect scene representations, potentially affecting the downstream task performance. 
In practice, redundancy across multiple orthographic projections mitigates partial occlusions.
Keyframe-based policies cover a broad range of tasks but are not suited for highly dynamic actions, such as object tossing or pressing with a desired force.
Computational cost is another consideration, as training requires substantial time and resources; approaches such as distillation or parameter-efficient fine-tuning provide clear avenues for improvement.
Markovian policy learning makes long-horizon tasks challenging, as multiple keyframes must be predicted. 
Finally, single-camera input may produce redundant or even noisy views, which can affect predictions; however, OG-VLA’s design ensures robustness to camera pose changes and allows future extensions with hierarchical or reactive controllers.
Together, these limitations define clear directions for future research while demonstrating that OG-VLA provides a practical, generalizable, and robust approach to 3D-aware VLAs for robotic manipulation.


\renewcommand*{\bibfont}{\small}
\printbibliography





\section*{Appendix}
\addcontentsline{toc}{chapter}{Appendix}
\renewcommand{\thesection}{\Alph{section}} 
\setcounter{section}{0}

\section{Tasks}

\begin{table}[h]
\centering
\noindent
\begin{minipage}{0.48\textwidth}
    \resizebox{\columnwidth}{!}{%
    \begin{tabular}{l l l c}
        \toprule
        \textbf{Task Types} & \textbf{Goal States} & \textbf{Success} & \textbf{Training} \\
        & & \textbf{Ranges} & \textbf{Data} \\
        \midrule
        Pickup Object & 10, 20, 30, 40 (cm) & $\pm$5 cm & 623 \\
        ReorientObject & 0, 45, 135, 180 (°) & $\pm$20° & 355 \\
        Open Drawer & 25, 50, 75, 100 (\%) & $\pm$10\% & 554 \\
        Close Drawer & 0, 25, 50, 75 (\%) & $\pm$10\% & 671 \\
        Open Cabinet & 25, 50, 75, 100 (\%) & $\pm$10\% & 319 \\
        Close Cabinet & 0, 25, 50, 75 (\%) & $\pm$10\% & 478 \\
        Pour Water & 25, 50, 75, 100 (\%) & $\pm$10\% & 312 \\
        Transfer Water & 20, 40, 60, 80 (\%) & $\pm$10\% & 259 \\
        \bottomrule
    \end{tabular}
    }
    \vspace{0.2cm}
    \caption{Overview of the 8 tasks in ARNOLD. Each task features 4
goal states specified by human language, one of which is reserved
for novel state evaluation and the other three are seen in the training dataset. The task is considered successful when the
object state remains in the success range for two seconds. For Transfer Water, an additional condition of only less than
10\% spillage of the original amount of water in the cup is imposed.}
\end{minipage}
\hfill
\begin{minipage}{0.48\textwidth}
    \centering
    \resizebox{\columnwidth}{!}{%
    \begin{tabular}{l l}
        \toprule
        \textbf{Tasks} & \textbf{Examples of Templates} \\
        \midrule
        Pickup Object & \textit{Raise} [value\_object] [value\_height] \textit{above the ground} \\
        Reorient Object & \textit{Reorient} [value\_object] [value\_degree] \textit{away from the up axis} \\
        Open Drawer & \textit{Open the} [value\_position] [value\_object] [value\_percent] \\
        Close Drawer & \textit{Close the} [value\_position] [value\_object] [value\_percent] \\
        Open Cabinet & \textit{Open the} [value\_position] [value\_object] [value\_percent] \\
        Close Cabinet & \textit{Close the} [value\_position] [value\_object] [value\_percent] \\
        Pour Water & \textit{Pour} [value\_percent] \textit{water out of} [value\_object] \\
        Transfer Water & \textit{Transfer} [value\_percent] \textit{water to} [value\_object] \\
        \bottomrule
    \end{tabular}
    }
    \vspace{0.2cm}
    \caption{
    Examples of instruction templates used for the tasks.
    }
\end{minipage}
\end{table}


\section{Prompts used in model variants}
\label{prompt-detail}
\subsection{\ours\ Prompt and Response}
\label{prompt-detail-ours}

\textbf{Prompt: }
Task: ``the bottle should be twenty centimeters from the ground.". Where should the robot move next? Show the gripper's next pose as translation and rotation heatmaps on the input orthographic views. Translation should be represented as red heatmap on all 4 views. Follow the provided instruction to compute correct translation points in the images. Rotation should be represented as yellow, blue, and green heatmaps for the front, top, and left views, corresponding to the x, z, and y axes respectively.

\textbf{Response: }
The next gripper pose for the given task 'the bottle should be twenty centimeters from the ground.' for timestep 1 is shown in the generated images.

\subsection{Text Action model Prompt and Response}
\label{prompt-detail-text}

\textbf{Prompt: }
Task: ``pull the top drawer 50\% open". Where should the robot move next? Format the robot's gripper action as a relative 3D coordinate, an Euler rotation, and a binary gripper open/close state. All numbers are floats with two decimal places, each in relative coordinates.

\textbf{Response: }
pos: [0.54, 0.42, 0.62], rot: [-1.57, -0.0, -1.57]

\section{Detailed Ablation Results and Analysis}
\label{detailed-ablations}

\subsection{Action Prediction Ablations}

We experimented with two common alternative architectural choices for action generation in VLAs, both of which failed to learn a working policy under similar training and evaluation settings as for \ours.

1) Text Action: is an architecture variant where the LLM also produces gripper's next pose in the form of text akin to prior VLA models~\cite{li2024llara}.
In general, the raw output sequence from the LLM can be of slightly different format and contain additional text (e.g. text like \emph{the next robot action is: }), so long as it contains the information above.
We apply regex parsing to extract from the text the gripper position $p^{text}$, orientation $\omega^{text}$.
The prompt used for this ablation is shown in Appendix~\ref{prompt-detail-text}.

Predicting actions as text without any visual reasoning on the output end of the model results in an imprecise action prediction model. 
\citet{li2024llara} have shown this architecture to work when training with large-scale robot datasets.
However, we find that with a small robotics dataset, it is hard for VLAs to learn precise control via direct text token prediction.

2) Action Tokens: is an architecture variant where the LLM produces special action tokens added to the LLM vocabulary, such as $[trans0]$ or $[rot0]$ for translation and rotation modalities, akin to that for the image modality.
This architecture is closer to works that perform action tokenization~\cite{kim24openvla}.
We decode the hidden state vectors for these tokens using additional MLP decoders to predict 3D translation and rotation vectors.

\begin{table*}[t]
\centering
\resizebox{\textwidth}{!}{%
\begin{tabular}{lcccccccccccccc}
\toprule
      & Pickup & Reorient & Open & Close & Open & Close & Pour & Transfer & Overall \\ 
Model & Object & Object & Drawer & Drawer & Cabinet & Cabinet & Water & Water &  \\ 
\midrule
\textbf{(1) No Reconstruction}   & 76.7 {\tiny$\pm$  6.2} & 10.0 {\tiny$\pm$ 10.8} &  3.3 {\tiny$\pm$  2.4} & 20.0 {\tiny$\pm$  0.0} &  1.7 {\tiny$\pm$  2.4} & 10.0 {\tiny$\pm$  7.1} & 28.3 {\tiny$\pm$ 11.8} & 13.3 {\tiny$\pm$ 12.5} & 20.4 {\tiny$\pm$  4.7} \\
\rowcolor{Gray}
\quad -Novel Object     & 56.7 {\tiny$\pm$  6.2} & 10.0 {\tiny$\pm$  0.0} &  3.3 {\tiny$\pm$  2.4} & 23.3 {\tiny$\pm$ 15.5} &  0.0 {\tiny$\pm$  0.0} & 10.0 {\tiny$\pm$  8.2} & 13.3 {\tiny$\pm$  8.5} &  6.7 {\tiny$\pm$  2.4} & 15.4 {\tiny$\pm$  5.1}
 \\
\rowcolor{Gray1}
\quad -Novel Scene    & 51.7 {\tiny$\pm$  8.5} &  6.7 {\tiny$\pm$  6.2} & 10.0 {\tiny$\pm$  4.1} & 10.0 {\tiny$\pm$  8.2} &  0.0 {\tiny$\pm$  0.0} &  5.0 {\tiny$\pm$  4.1} & 11.7 {\tiny$\pm$  6.2} &  6.7 {\tiny$\pm$  6.2} & 12.7 {\tiny$\pm$  3.0}

 \\
\rowcolor{Gray2}
\quad -Novel State     &  0.0 {\tiny$\pm$  0.0} &  6.7 {\tiny$\pm$  6.2} &  6.7 {\tiny$\pm$  2.4} &  5.0 {\tiny$\pm$  0.0} &  0.0 {\tiny$\pm$  0.0} &  0.0 {\tiny$\pm$  0.0} & 13.3 {\tiny$\pm$  8.5} &  0.0 {\tiny$\pm$  0.0} &  4.0 {\tiny$\pm$  2.1}
 \\
\midrule
\textbf{(2) Reconstruction} & 86.7 {\tiny$\pm$  2.4} & 15.0 {\tiny$\pm$  7.1} & 38.3 {\tiny$\pm$  2.4} & 51.7 {\tiny$\pm$  2.4} &  0.0 {\tiny$\pm$  0.0} & 16.7 {\tiny$\pm$  2.4} & 25.0 {\tiny$\pm$  4.1} & 16.7 {\tiny$\pm$  6.2} & 31.2 {\tiny$\pm$  2.3}
 \\
\rowcolor{Gray}
\quad -Novel Object        & 85.0 {\tiny$\pm$  4.1} &  0.0 {\tiny$\pm$  0.0} &  1.7 {\tiny$\pm$  2.4} & 55.0 {\tiny$\pm$ 10.8} &  0.0 {\tiny$\pm$  0.0} &  5.0 {\tiny$\pm$  4.1} & 18.3 {\tiny$\pm$  2.4} &  6.7 {\tiny$\pm$  6.2} & 21.5 {\tiny$\pm$  0.3}
 \\
\rowcolor{Gray1}
\quad -Novel Scene         & 73.3 {\tiny$\pm$  2.4} &  1.7 {\tiny$\pm$  2.4} & 26.7 {\tiny$\pm$  9.4} & 36.7 {\tiny$\pm$  4.7} &  1.7 {\tiny$\pm$  2.4} &  1.7 {\tiny$\pm$  2.4} & 16.7 {\tiny$\pm$  9.4} &  8.3 {\tiny$\pm$  2.4} & 20.8 {\tiny$\pm$  1.1}
 \\
\rowcolor{Gray2}
\quad -Novel State         &  0.0 {\tiny$\pm$  0.0} & 13.3 {\tiny$\pm$  6.2} & 13.3 {\tiny$\pm$  2.4} & 20.0 {\tiny$\pm$  0.0} &  0.0 {\tiny$\pm$  0.0} &  0.0 {\tiny$\pm$  0.0} &  8.3 {\tiny$\pm$  6.2} & 13.3 {\tiny$\pm$  2.4} &  8.5 {\tiny$\pm$  1.6}
 \\
\midrule
\textbf{(3) Faded Reconstruction}   & 93.3 {\tiny$\pm$  2.4} &  5.0 {\tiny$\pm$  4.1} & 23.3 {\tiny$\pm$  2.4} & 36.7 {\tiny$\pm$  8.5} &  0.0 {\tiny$\pm$  0.0} &  5.0 {\tiny$\pm$  4.1} & 30.0 {\tiny$\pm$  0.0} & 20.0 {\tiny$\pm$ 10.8} & 26.7 {\tiny$\pm$  2.3} \\
\rowcolor{Gray}
\quad -Novel Object        & 75.0 {\tiny$\pm$  4.1} & 18.3 {\tiny$\pm$  2.4} &  0.0 {\tiny$\pm$  0.0} & 55.0 {\tiny$\pm$  4.1} &  0.0 {\tiny$\pm$  0.0} &  8.3 {\tiny$\pm$  4.7} & 36.7 {\tiny$\pm$ 11.8} &  5.0 {\tiny$\pm$  4.1} & 24.8 {\tiny$\pm$  2.3} \\
\rowcolor{Gray1}
\quad -Novel Scene         & 76.7 {\tiny$\pm$  6.2} & 11.7 {\tiny$\pm$  6.2} & 28.3 {\tiny$\pm$  4.7} & 28.3 {\tiny$\pm$  2.4} &  5.0 {\tiny$\pm$  4.1} &  8.3 {\tiny$\pm$  2.4} & 20.0 {\tiny$\pm$  8.2} & 11.7 {\tiny$\pm$  6.2} & 23.8 {\tiny$\pm$  3.1} \\
\rowcolor{Gray2}
\quad -Novel State         &  0.0 {\tiny$\pm$  0.0} & 18.3 {\tiny$\pm$ 10.3} &  6.7 {\tiny$\pm$  2.4} & 10.0 {\tiny$\pm$  4.1} &  1.7 {\tiny$\pm$  2.4} &  0.0 {\tiny$\pm$  0.0} &  1.7 {\tiny$\pm$  2.4} & 11.7 {\tiny$\pm$  6.2} &  6.2 {\tiny$\pm$  0.5} \\
\bottomrule
\end{tabular}
}
\caption{\textbf{Success rates for image generation modes for each task.} We show that generating actions with reconstruction or with faded reconstructions work better than that without reconstruction.}

\label{tab:results2-full}
\end{table*}

\begin{table*}[h!]
\centering
\resizebox{\textwidth}{!}{%
\begin{tabular}{lcccccccccccccc}
\toprule
      & Pickup & Reorient & Open & Close & Open & Close & Pour & Transfer & Overall \\ 
Model & Object & Object & Drawer & Drawer & Cabinet & Cabinet & Water & Water &  \\ 
\midrule
\textbf{\ours{}} & 86.7 {\tiny$\pm$  2.4} & 15.0 {\tiny$\pm$  7.1} & 38.3 {\tiny$\pm$  2.4} & 51.7 {\tiny$\pm$  2.4} &  0.0 {\tiny$\pm$  0.0} & 16.7 {\tiny$\pm$  2.4} & 25.0 {\tiny$\pm$  4.1} & 16.7 {\tiny$\pm$  6.2} & 31.2 {\tiny$\pm$  2.3}
 \\
\rowcolor{Gray}
\quad -Novel Object        & 85.0 {\tiny$\pm$  4.1} &  0.0 {\tiny$\pm$  0.0} &  1.7 {\tiny$\pm$  2.4} & 55.0 {\tiny$\pm$ 10.8} &  0.0 {\tiny$\pm$  0.0} &  5.0 {\tiny$\pm$  4.1} & 18.3 {\tiny$\pm$  2.4} &  6.7 {\tiny$\pm$  6.2} & 21.5 {\tiny$\pm$  0.3}
 \\
\rowcolor{Gray1}
\quad -Novel Scene         & 73.3 {\tiny$\pm$  2.4} &  1.7 {\tiny$\pm$  2.4} & 26.7 {\tiny$\pm$  9.4} & 36.7 {\tiny$\pm$  4.7} &  1.7 {\tiny$\pm$  2.4} &  1.7 {\tiny$\pm$  2.4} & 16.7 {\tiny$\pm$  9.4} &  8.3 {\tiny$\pm$  2.4} & 20.8 {\tiny$\pm$  1.1}
 \\
\rowcolor{Gray2}
\quad -Novel State         &  0.0 {\tiny$\pm$  0.0} & 13.3 {\tiny$\pm$  6.2} & 13.3 {\tiny$\pm$  2.4} & 20.0 {\tiny$\pm$  0.0} &  0.0 {\tiny$\pm$  0.0} &  0.0 {\tiny$\pm$  0.0} &  8.3 {\tiny$\pm$  6.2} & 13.3 {\tiny$\pm$  2.4} &  8.5 {\tiny$\pm$  1.6}
 \\
\midrule
\textbf{+Tiled Views}  & 75.0 {\tiny$\pm$  7.1} &  6.7 {\tiny$\pm$  4.7} & 33.3 {\tiny$\pm$  2.4} & 28.3 {\tiny$\pm$  6.2} &  1.7 {\tiny$\pm$  2.4} & 20.0 {\tiny$\pm$  4.1} & 15.0 {\tiny$\pm$ 10.8} & 18.3 {\tiny$\pm$  6.2} & 24.8 {\tiny$\pm$  0.3}
 \\
\rowcolor{Gray}
\quad -Novel Object        & 58.3 {\tiny$\pm$  8.5} & 15.0 {\tiny$\pm$  7.1} &  1.7 {\tiny$\pm$  2.4} & 30.0 {\tiny$\pm$  0.0} &  0.0 {\tiny$\pm$  0.0} & 10.0 {\tiny$\pm$  4.1} & 40.0 {\tiny$\pm$  4.1} &  6.7 {\tiny$\pm$  6.2} & 20.2 {\tiny$\pm$  1.3}
 \\
\rowcolor{Gray1}
\quad -Novel Scene         & 60.0 {\tiny$\pm$  7.1} & 15.0 {\tiny$\pm$ 10.8} & 45.0 {\tiny$\pm$  7.1} & 21.7 {\tiny$\pm$  6.2} &  5.0 {\tiny$\pm$  4.1} &  5.0 {\tiny$\pm$  4.1} & 26.7 {\tiny$\pm$  6.2} &  1.7 {\tiny$\pm$  2.4} & 22.5 {\tiny$\pm$  0.5}
 \\
\rowcolor{Gray2}
\quad -Novel State         &  0.0 {\tiny$\pm$  0.0} & 10.0 {\tiny$\pm$  4.1} & 10.0 {\tiny$\pm$  7.1} & 18.3 {\tiny$\pm$  4.7} &  1.7 {\tiny$\pm$  2.4} &  1.7 {\tiny$\pm$  2.4} &  1.7 {\tiny$\pm$  2.4} & 16.7 {\tiny$\pm$ 11.8} &  7.5 {\tiny$\pm$  2.6}
 \\
\midrule


\textbf{-LLM}  & 86.7 {\tiny$\pm$  2.4} &  5.0 {\tiny$\pm$  7.1} &  6.7 {\tiny$\pm$  4.7} & 33.3 {\tiny$\pm$  4.7} &  0.0 {\tiny$\pm$  0.0} &  6.7 {\tiny$\pm$  4.7} & 15.0 {\tiny$\pm$  0.0} &  6.7 {\tiny$\pm$  2.4} & 20.0 {\tiny$\pm$  1.5}
 \\
\rowcolor{Gray}
\quad -Novel Object        & 68.3 {\tiny$\pm$  6.2} &  1.7 {\tiny$\pm$  2.4} &  6.7 {\tiny$\pm$  9.4} & 40.0 {\tiny$\pm$  4.1} &  0.0 {\tiny$\pm$  0.0} &  3.3 {\tiny$\pm$  2.4} & 10.0 {\tiny$\pm$  4.1} &  8.3 {\tiny$\pm$  2.4} & 17.3 {\tiny$\pm$  1.6}
 \\
\rowcolor{Gray1}
\quad -Novel Scene         & 71.7 {\tiny$\pm$  6.2} &  8.3 {\tiny$\pm$  6.2} & 18.3 {\tiny$\pm$  2.4} & 21.7 {\tiny$\pm$  8.5} &  3.3 {\tiny$\pm$  2.4} &  0.0 {\tiny$\pm$  0.0} & 15.0 {\tiny$\pm$  4.1} &  5.0 {\tiny$\pm$  4.1} & 17.9 {\tiny$\pm$  3.3}
 \\
\rowcolor{Gray2}
\quad -Novel State        &  0.0 {\tiny$\pm$  0.0} &  0.0 {\tiny$\pm$  0.0} &  6.7 {\tiny$\pm$  2.4} & 16.7 {\tiny$\pm$  2.4} &  0.0 {\tiny$\pm$  0.0} &  1.7 {\tiny$\pm$  2.4} &  3.3 {\tiny$\pm$  2.4} & 10.0 {\tiny$\pm$  4.1} &  4.8 {\tiny$\pm$  0.8}
 \\
\midrule

\textbf{+Tiled Views -LLM} & 71.7 {\tiny$\pm$ 10.3} &  1.7 {\tiny$\pm$  2.4} & 13.3 {\tiny$\pm$  8.5} & 16.7 {\tiny$\pm$  4.7} &  0.0 {\tiny$\pm$  0.0} &  8.3 {\tiny$\pm$  2.4} & 15.0 {\tiny$\pm$  8.2} & 10.0 {\tiny$\pm$  0.0} & 17.1 {\tiny$\pm$  3.4}
 \\
\rowcolor{Gray}
\quad -Novel Object        & 56.7 {\tiny$\pm$  8.5} &  8.3 {\tiny$\pm$  2.4} &  1.7 {\tiny$\pm$  2.4} & 16.7 {\tiny$\pm$  8.5} &  0.0 {\tiny$\pm$  0.0} &  1.7 {\tiny$\pm$  2.4} & 15.0 {\tiny$\pm$  4.1} &  6.7 {\tiny$\pm$  2.4} & 13.3 {\tiny$\pm$  1.6}
 \\
\rowcolor{Gray1}
\quad -Novel Scene         & 61.7 {\tiny$\pm$  6.2} &  5.0 {\tiny$\pm$  4.1} & 20.0 {\tiny$\pm$  4.1} & 11.7 {\tiny$\pm$  6.2} &  0.0 {\tiny$\pm$  0.0} &  3.3 {\tiny$\pm$  4.7} & 10.0 {\tiny$\pm$  0.0} & 10.0 {\tiny$\pm$  0.0} & 15.2 {\tiny$\pm$  1.2}
 \\
\rowcolor{Gray2}
\quad -Novel State        &  0.0 {\tiny$\pm$  0.0} &  6.7 {\tiny$\pm$  2.4} & 10.0 {\tiny$\pm$  0.0} & 30.0 {\tiny$\pm$  4.1} &  1.7 {\tiny$\pm$  2.4} &  0.0 {\tiny$\pm$  0.0} &  6.7 {\tiny$\pm$  6.2} &  3.3 {\tiny$\pm$  4.7} &  7.3 {\tiny$\pm$  0.8}
 \\
\midrule

\textbf{-Instruction to \textsc{IG}}  & 71.7 {\tiny$\pm$  4.7} &  8.3 {\tiny$\pm$  2.4} & 20.0 {\tiny$\pm$ 10.8} & 40.0 {\tiny$\pm$ 12.2} &  1.7 {\tiny$\pm$  2.4} & 11.7 {\tiny$\pm$  9.4} &  5.0 {\tiny$\pm$  4.1} & 15.0 {\tiny$\pm$  4.1} & 21.7 {\tiny$\pm$  2.6}
 \\
\rowcolor{Gray}
\quad -Novel Object        & 66.7 {\tiny$\pm$  6.2} &  0.0 {\tiny$\pm$  0.0} &  1.7 {\tiny$\pm$  2.4} & 45.0 {\tiny$\pm$  4.1} &  0.0 {\tiny$\pm$  0.0} &  8.3 {\tiny$\pm$  2.4} & 20.0 {\tiny$\pm$  4.1} &  1.7 {\tiny$\pm$  2.4} & 17.9 {\tiny$\pm$  0.8}
 \\
\rowcolor{Gray1}
\quad -Novel Scene         & 50.0 {\tiny$\pm$  8.2} & 11.7 {\tiny$\pm$  2.4} & 25.0 {\tiny$\pm$  0.0} & 26.7 {\tiny$\pm$  2.4} & 10.0 {\tiny$\pm$  0.0} & 11.7 {\tiny$\pm$  2.4} & 13.3 {\tiny$\pm$  4.7} &  5.0 {\tiny$\pm$  4.1} & 19.2 {\tiny$\pm$  2.1}
 \\
\rowcolor{Gray2}
\quad -Novel State         &  0.0 {\tiny$\pm$  0.0} &  3.3 {\tiny$\pm$  4.7} &  8.3 {\tiny$\pm$  6.2} & 13.3 {\tiny$\pm$  8.5} &  0.0 {\tiny$\pm$  0.0} &  0.0 {\tiny$\pm$  0.0} &  1.7 {\tiny$\pm$  2.4} &  8.3 {\tiny$\pm$  2.4} &  4.4 {\tiny$\pm$  1.8}
 \\
\bottomrule
\end{tabular}
}
\caption{\textbf{Model ablation results for each task.} We ablate components of \ours\ to justify our design choices such as using tiled views, and the contribution of the LLM in the pipeline.}

\label{tab:results3-full}
\end{table*}

The model still struggles to predict sufficiently precise actions to perform the tasks. 
This may be due to insufficient visual reasoning available for action decoding, as the decoders use the hidden states corresponding to the special token produced by the LLM. 
We also observed degeneration of LLM's ability to produce coherent and grammatical text as the training progressed for this model design.
This may have been caused by the training of new tokens and additional decoders failing to preserve original reasoning capabilities of the model, an issue that might be addressed by co-training with the pretraining datasets.
We do not observe degeneration when finetuning the model with original architectural components in \ours\ without any co-training with other datasets.
We leave the study with co-training for above architectural variants to future works.

\subsection{Image Generation Mode}


We study three image generation modes for action prediction:
(1) Generation without reconstruction: an all black image background
(2) Generation with reconstruction: an RGB image that is a reconstruction of the input image, and 
(3) Generation with faded reconstruction: a shifted RGB image between the range $[0, 128]$ that is a reconstruction of the input image.
For each image mode, the action Gaussian distributions are overlaid on these backgrounds.
These three choices carry tradeoffs.
The first method is the purest way to represent actions, but it appears challenging for image generators pre-trained on generating color images to learn to generate uniform black backgrounds. 
The second method does not require un-learning generation of color images, but burdens the generator with the additional reconstruction task, which has the potential to take model capacity away from action generation. 
On the flipside, it has the potential for some positive cross-task transfer, and better scene understanding and visual reasoning. 
The third method is a middle ground between the other two methods, which we include in our study. It eases action decoding by keeping values in the $(128, 255]$ range exclusively for action annotations. 
For methods (2) and (3), we apply an additional filtering step to identify the Gaussian and recover a grayscale heatmap.

\begin{figure*}[t!]
    \centering
    \includegraphics[width=\textwidth]{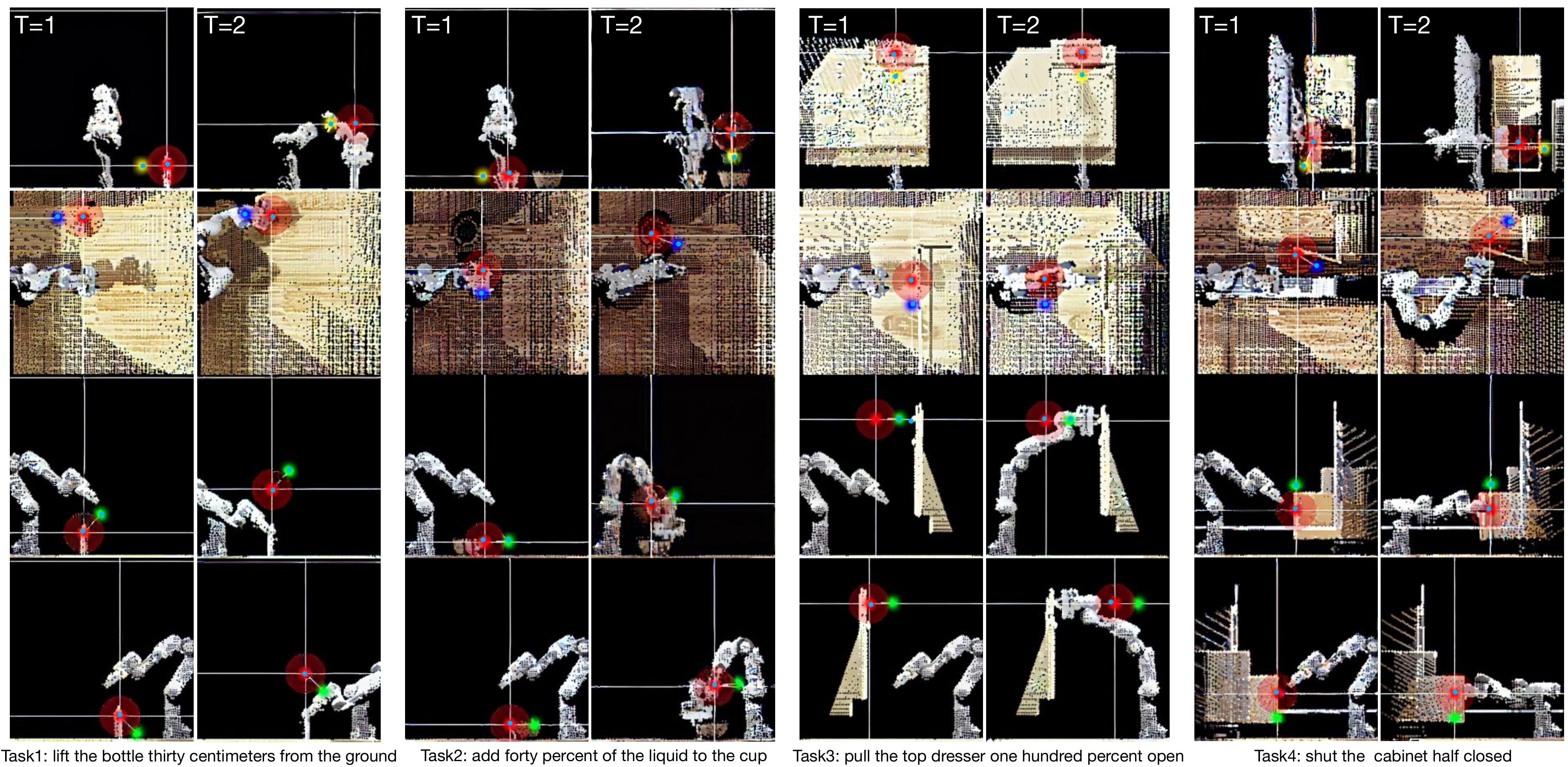}
    \includegraphics[width=\textwidth]{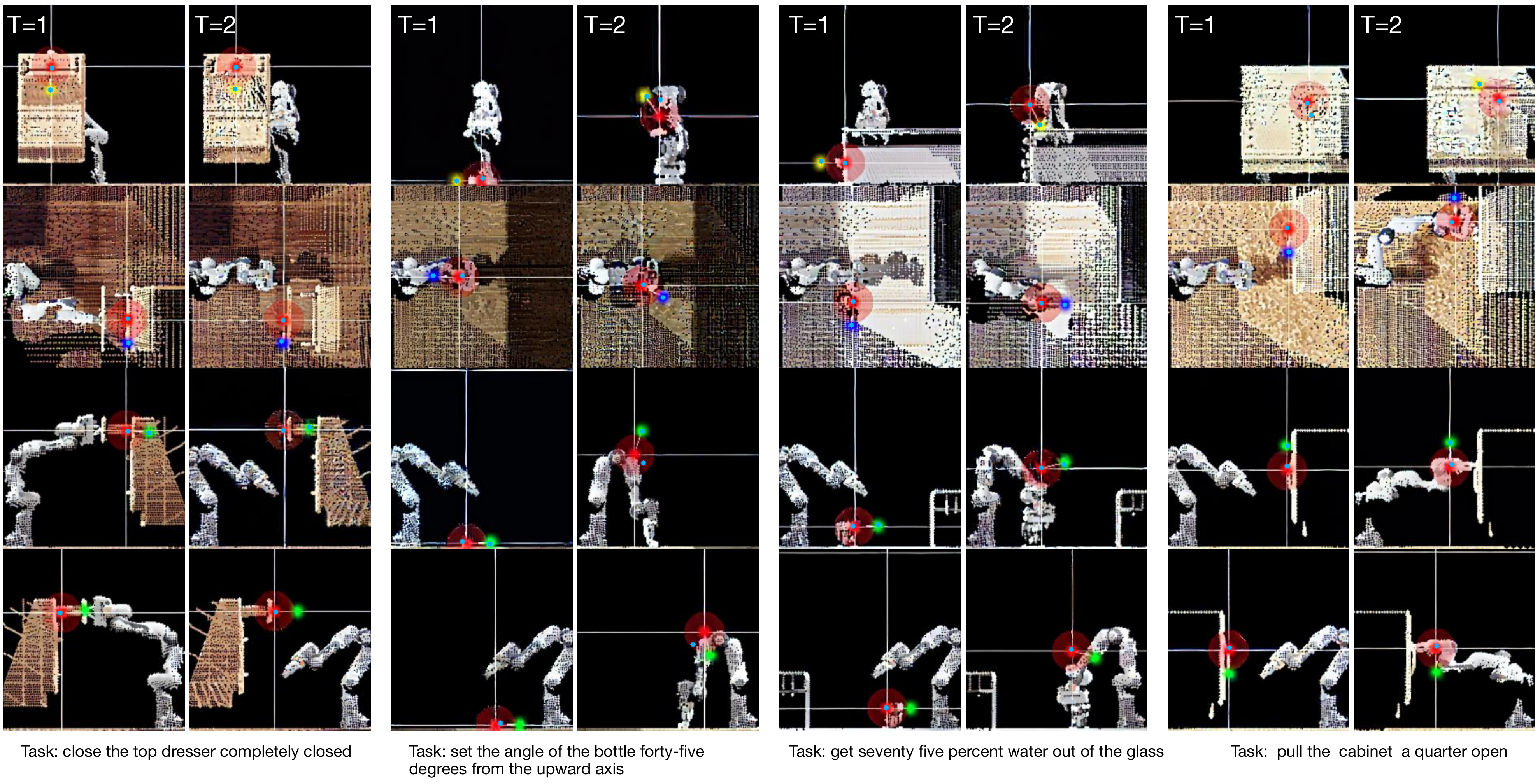}
    \caption{\textbf{Example gripper position and rotation outputs} from \ours\ for eight different tasks. The rows are the different views: front, top, left, and right. For each task, the two columns are two timesteps required to solve the task. The red Gaussian is the predicted position. The yellow, blue, and green Gaussians are predicted rotation angles along x, z, and y-axis respectively. The blue dot is our model's output gripper position, back-projected to each view. The dots on rotation Gaussians are showing the extracted pixel for computing the rotation angle in reference to the horizontal right axis in each view.}
    \label{fig:examples}
\end{figure*}

\begin{table*}[t]
\centering
\resizebox{\textwidth}{!}{%
\begin{tabular}{@{}lcccccccccccccccc@{}}
\toprule
   & Pickup & Reorient & Open & Close & Open & Close & Pour & Transfer & Overall \\ 
                    & Object & Object & Drawer & Drawer & Cabinet & Cabinet & Water & Water &  \\ 
\midrule
\rowcolor{Gray1}
Training set evaluation            & 76.7 {\tiny$\pm$  4.7} & 11.7 {\tiny$\pm$  2.4} & 35.0 {\tiny$\pm$  0.0} & 58.3 {\tiny$\pm$  4.7} &  0.0 {\tiny$\pm$  0.0} & 10.0 {\tiny$\pm$  4.1} & 18.3 {\tiny$\pm$  6.2} & 20.0 {\tiny$\pm$  4.1} & 28.8 {\tiny$\pm$  0.5}  \\
+ Ground Truth Translation        & 91.7 {\tiny$\pm$  2.4} & 21.7 {\tiny$\pm$ 14.3} & 76.7 {\tiny$\pm$  6.2} & 81.7 {\tiny$\pm$  2.4} & 15.0 {\tiny$\pm$  7.1} & 21.7 {\tiny$\pm$  4.7} & 28.3 {\tiny$\pm$  2.4} & 33.3 {\tiny$\pm$  2.4} & 46.2 {\tiny$\pm$  0.9}  \\
\rowcolor{Gray1}
+ Ground Truth Rotation          & 96.7 {\tiny$\pm$  4.7} & 60.0 {\tiny$\pm$  4.1} & 43.3 {\tiny$\pm$  6.2} & 75.0 {\tiny$\pm$  8.2} &  5.0 {\tiny$\pm$  0.0} & 40.0 {\tiny$\pm$  7.1} & 40.0 {\tiny$\pm$  8.2} & 11.7 {\tiny$\pm$  2.4} & 46.5 {\tiny$\pm$  3.8} \\
\bottomrule
\end{tabular}
}
\caption{\textbf{Ablated evaluation of the model's translation and rotation prediction capabilities} on a sampled training set of 20 episodes, similar to the test sets. In each evaluation, we individually ablate either the translation or rotation predictions to assess the model's prediction capabilities for each.}
\label{tab:results4}
\end{table*}

We report success rates across tasks and test splits in Table~\ref{tab:results2-full}.
Generation without reconstruction of the scene performs the worst of all generation modes. 
This may be due to forcing the \textsc{ImageGenerator} to unlearn it's prior of generating natural images and forcing it to only predict Gaussian distributions.
We also observe instability in training this version, as sometimes the \textsc{ImageGenerator} would collapse to produce completely black or noisy images and stop generating actions on them.
Generation with reconstruction and that with faded reconstruction did not show consistent difference over all test splits. 
Generation with reconstruction learns a policy that's better by 4.4\% on Novel Pose split and 2.3\% Novel State split. 
Generation with faded reconstruction performs better on Novel Object and Novel State splits by 3.3\% and 3\%.
Therefore, we conclude that these model variants have similar performance, and generation with faded reconstruction did not work better as we had hypothesized due to its balance between forcing the model to focus less on scene reconstruction and more on action prediction.
Therefore, we report generation with reconstruction as our final method.

\begin{figure*}[h!]
    \centering
    \begin{subfigure}[t]{0.32\textwidth}
        \centering
        \includegraphics[width=\textwidth]{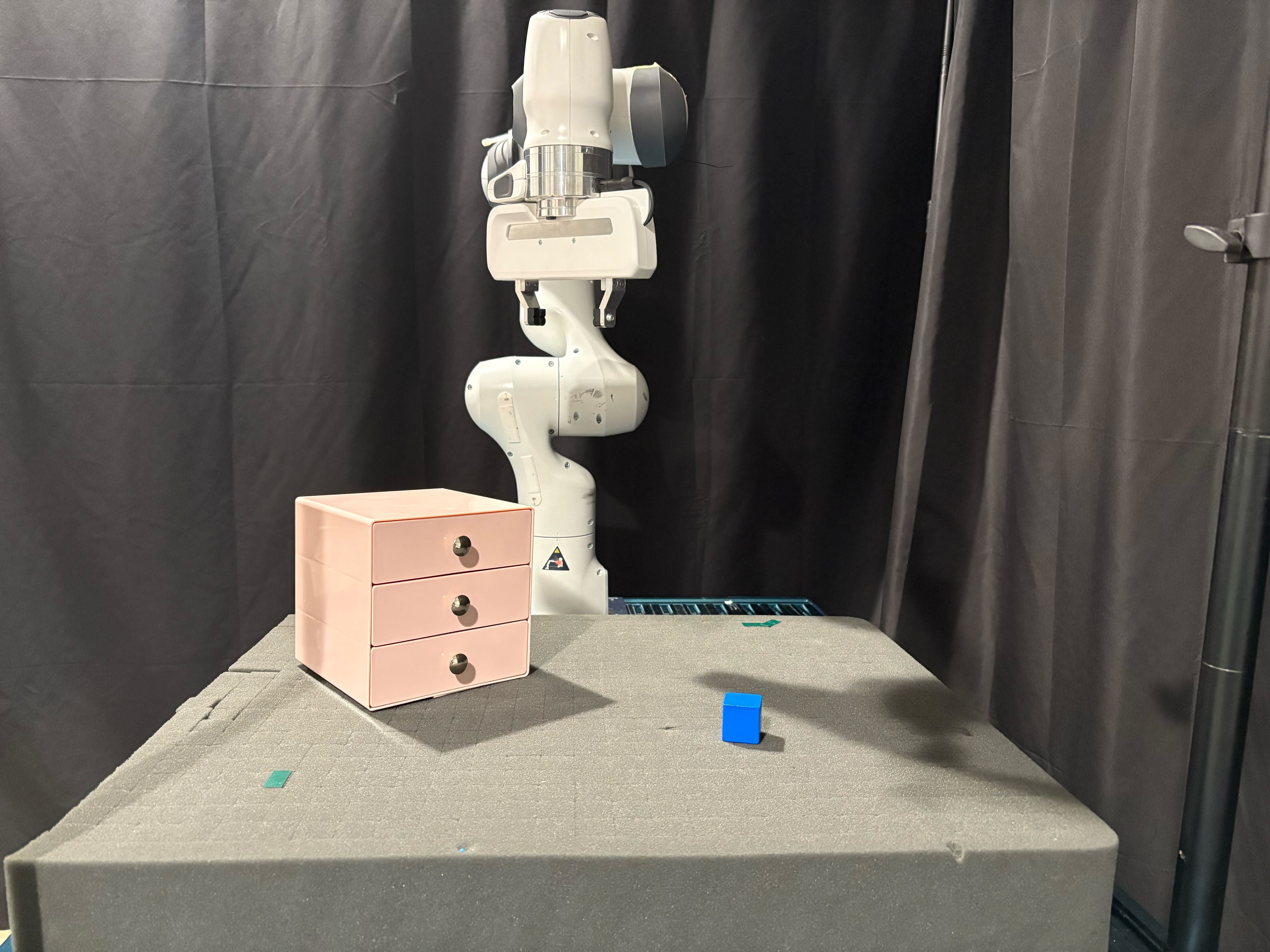}
        \caption{Training demonstration objects}
        \label{fig:sub1}
    \end{subfigure}
    \hfill
    \begin{subfigure}[t]{0.32\textwidth}
        \centering
        \includegraphics[width=\textwidth]{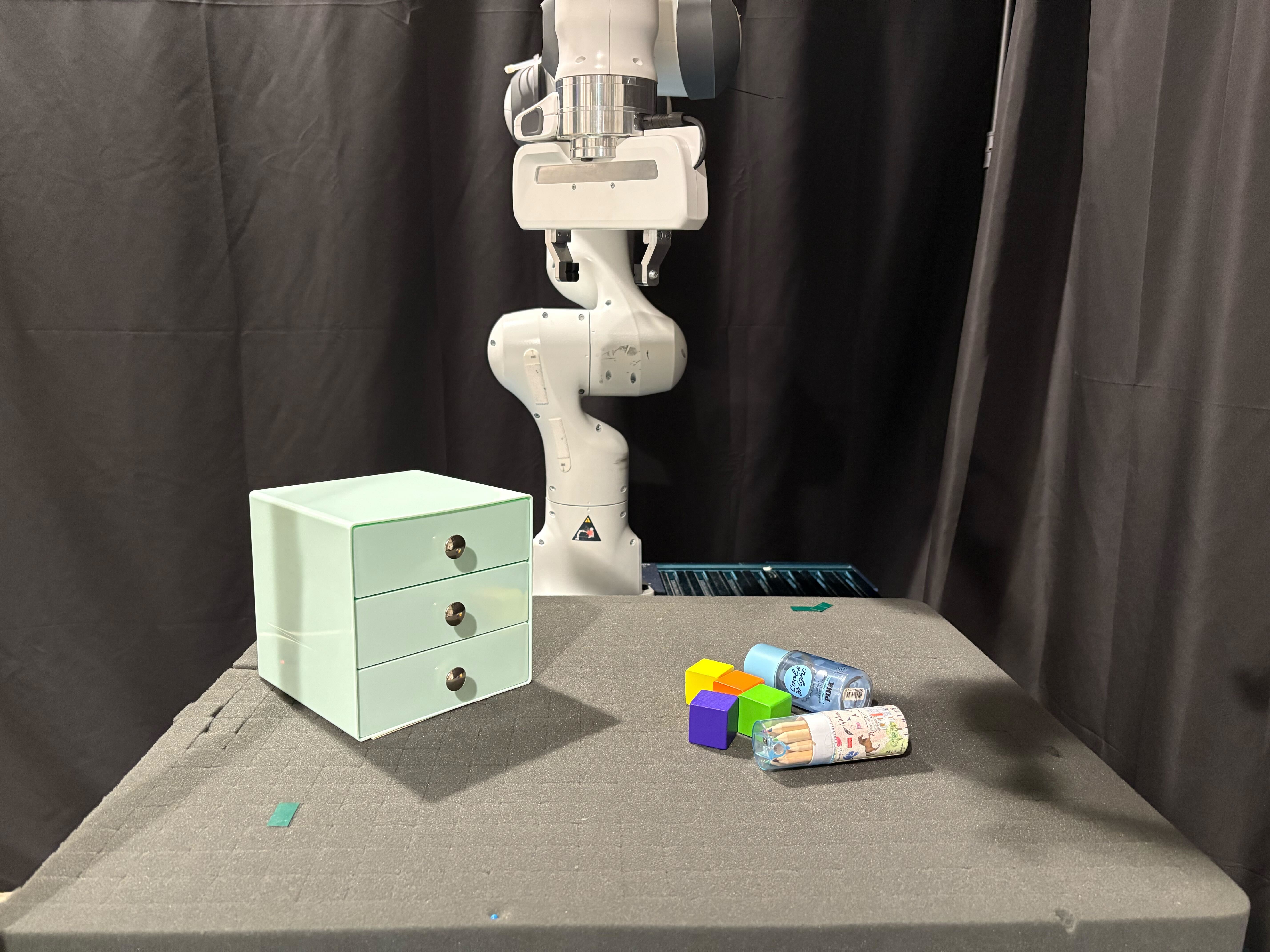}
        \caption{Novel objects used at test time}
        \label{fig:sub2}
    \end{subfigure}
    \hfill
    \begin{subfigure}[t]{0.32\textwidth}
        \centering
    \includegraphics[width=\textwidth]{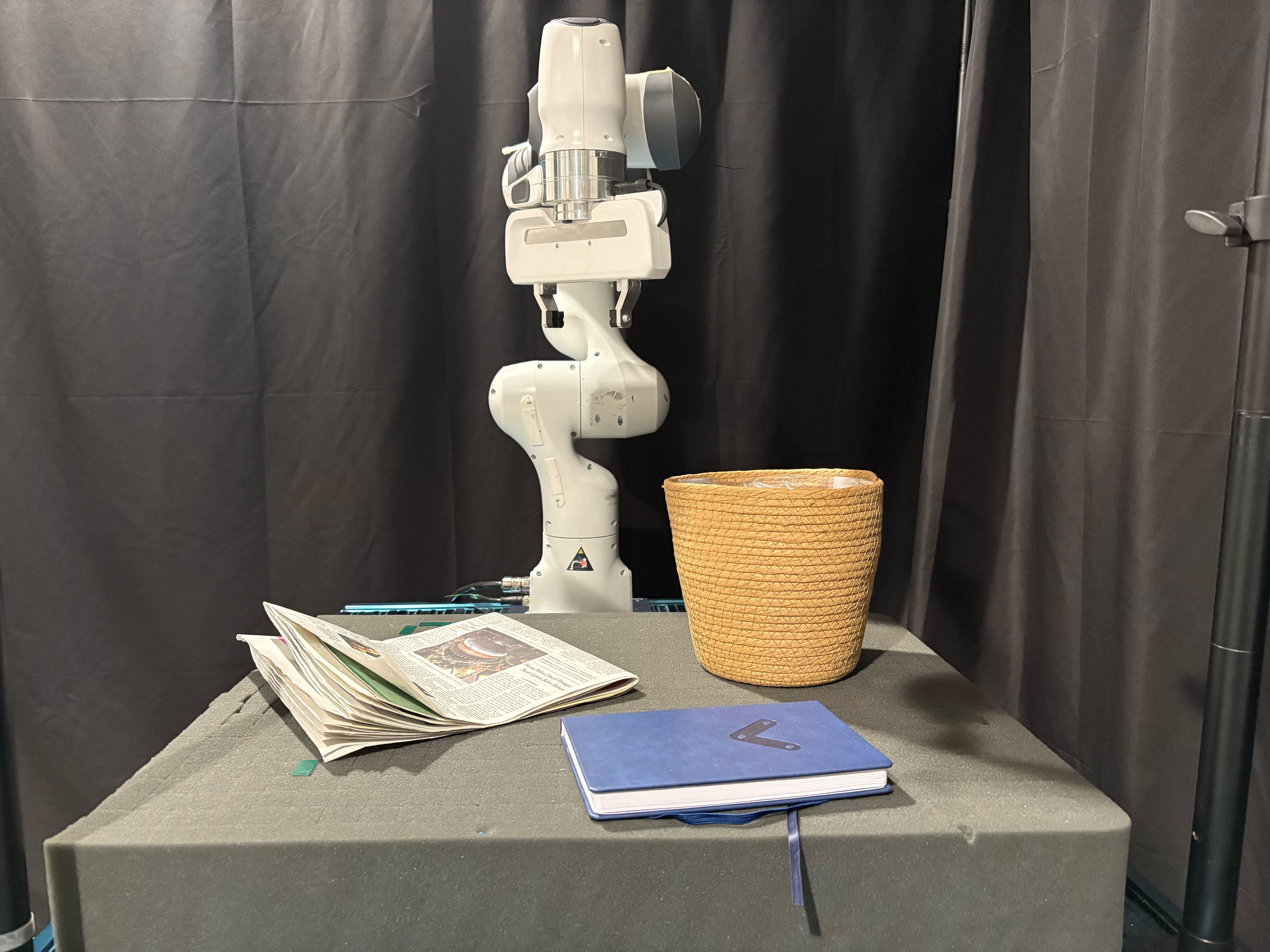}
        \caption{Novel scene distractors used at test time}
        \label{fig:sub3}
    \end{subfigure}

    \caption{Real-world setup: objects used at training and test-time. For novel scene, we additional vary the lighting and background by switching on/off external lighting source, and removing curtains.}
    \label{fig:real-world-setup}
\end{figure*}

\subsection{Model Ablations}

We present ablations on \ours's design choices in Table~\ref{tab:results3-full}. 
The first result shows the effect of tiling the 4 orthographic views instead of feeding them to the \textsc{ImageGenerator} in batch of 4 as also explored in Genima~\cite{shridhar2024generative}.
For tiling, we stack the 4 views as in 2D array, following prior work.
Tiling did not improve the performance for our model as opposed to results reported in the prior works. 
This may have occurred because the prior work did not have an LLM in the pipeline, so tiling becomes necessary for modeling interactions between views to generate multi-view consistent predictions.
However, due to the LLM in \ours conditioning generation in all views, there's sufficient interaction and reasoning between views through the predicted image tokens $\left(t^{i}_{a},\quad i \in \{1,\dots,4\}\right)$.

The second experiment ablates the LLM to study its contribution to the overall performance.
We remove the LLM from the system, directly passing image and text representations to the $\textsc{ImageGenerator}$.
This result shows a drop in performance, highlighting the importance of the LLM in our pipeline. 

Next, we study adding tiling to the previous LLM ablation to ensure that interaction between views, which was earlier happening through the LLM, can now take place in the $\textsc{ImageGenerator}$.
Tiling further leads to a drop in performance, perhaps because of the reduction in the number of tokens used to represent each view. 
Without tiling, each image is represented by 256 patch tokens, but with tiling all 4 images are represented by 256 tokens in total, potentially creating too tight a representational bottleneck.
This version is also similar to Genima with only an \textsc{ImageGenerator} and tiled camera input views in the pipeline.

Finally, we study the effect of bypassing the prompt and instruction into the \textsc{ImageGenerator}, shown in the last section of Table~\ref{tab:results3-full}. 
The performance drop suggests that some instruction information may have been lost in image tokens output from the LLM, therefore it is beneficial to provide that information separately to the $\textsc{ImageGenerator}$ for more accurate action prediction.

\subsection{Qualitative results}
We present qualitative results in Figure~\ref{fig:examples} of the output of \ours, showing the generated actions, inferred gripper positions and rotations.
We show predictions for the tasks: Pickup Object, Transfer Water, Open Drawer, and Close Cabinet. We add remaining task prediction examples in the Appendix. 
These visualizations show that \ours\ can do complex translation and rotation tasks, that require both object and free space reasoning.
We observe that the translation predictions are quite consistent for most cases, except for Task3 at T=1. 
The prediction in the Left view is incorrect, however, due to correct predictions in other views, the most likely 3D point (blue dot) is still correctly extracted at the handle of the drawer using the optimization in Equation~\ref{eq:argmax3d}.
In Task4, we show a failure case where the task is to half-close the cabinet, however the predicted point in T=2 is less than the half-way point.

\subsection{Ablation with ground truth translation and rotation}


We present another set of ablations in Table~\ref{tab:results4} for studying our model's translation and rotation prediction abilities in an ablated setting on a sampled training set. 
First, we note that the overall performance of our model on Training and Novel Pose split is quite similar, which indicates that the model has not overfit and is generalizing well w.r.t the training performance. 
We observe that there is huge scope for improvement, both in predicting more precise translations and rotations from the last two rows of the table.
We believe that this gap can be closed with training techniques like co-training with other datasets for better reasoning~\cite{yuan2024robopoint} and leveraging existing large robotics datasets like that in other VLA models~\cite{kim24openvla}. 
We believe leveraging these datasets with a 3D-aware VLA framework can unlock better learning signals from these large-scale datasets and significantly improve results for our proposed VLA architecture.

\section{Real world details}
\label{real-world-details}
We calibrate the camera using MoveIt hand-eye calibration using an ArUco board~\footnote{\url{https://github.com/moveit/moveit_calibration}}.
We record the trajectory at 30 Hz frame rate. For training \ours, we only use the first 1/5-th of the trajectory before each annotated keyframe action in the trajectory, for augmenting states prior to each keyframe. 

Figure~\ref{fig:real-world-setup} shows the different training and test scenes. We provide evaluation videos of successful and failed trajectories for these test scenes on our website.




\end{document}